\pdfoutput=1

\documentclass[11pt]{article}

\usepackage{acl}

\usepackage{times}
\usepackage{latexsym}

\usepackage[T1]{fontenc}

\usepackage[utf8]{inputenc}

\usepackage{microtype}

\usepackage{inconsolata}

\usepackage{graphicx}
\usepackage{algorithmic}
\usepackage{algorithm}

\usepackage{CJKutf8}
\usepackage{amsmath}
\usepackage{subcaption}
\usepackage{amsfonts}
\usepackage{multirow}
\usepackage{booktabs}
\usepackage{colortbl}
\newcommand{\method}{SeqPO-SiMT}
\newcommand{\methodb}{SeqPO-SiMT }
\newcommand{\methoda}{SeqPO-SiMT}
\definecolor{mygreen}{RGB}{0, 128, 0}
\definecolor{lightblue}{RGB}{212, 221, 238}
\definecolor{mydarkblue}{RGB}{161,182,220}
\definecolor{lightorange}{RGB}{250, 224, 207}
\definecolor{darkorange}{RGB}{246, 188, 151}

\usepackage{calc}

\newlength{\DepthReference}
\newlength{\HeightReference}
\newlength{\Width}%

\newcommand{\MyColorBox}[2][red]%
{%
    \settowidth{\Width}{#2}%
    \setlength{\fboxsep}{0pt}%
    \colorbox{#1}%
    {%
        \raisebox{-\DepthReference}%
        {%
                \parbox[b][\HeightReference+\DepthReference][c]{\Width}{\centering#2}%
        }%
    }%
}

\def\1{\bm{1}}

\usepackage{bm}

\def\rvg{{\mathbf{g}}}

\def\rvx{{\mathbf{x}}}
\def\rvy{{\mathbf{y}}}

\DeclareMathAlphabet{\mathsfit}{\encodingdefault}{\sfdefault}{m}{sl}
\SetMathAlphabet{\mathsfit}{bold}{\encodingdefault}{\sfdefault}{bx}{n}

\title{SeqPO-SiMT: Sequential Policy Optimization for \\ Simultaneous Machine Translation}

\author{
 \textbf{Ting Xu\textsuperscript{\rm $\spadesuit$}}\footnotemark[1],
 \textbf{Zhichao Huang\textsuperscript{\rm $\clubsuit$}}\footnotemark[2],
 \textbf{Jiankai Sun\textsuperscript{\rm $\diamondsuit$}},
 \textbf{Shanbo Cheng\textsuperscript{\rm $\clubsuit$}}\footnotemark[2],
 \textbf{Wai Lam\textsuperscript{\rm $\spadesuit$}},
\\
 \textsuperscript{\rm $\spadesuit$}The Chinese University of Hong Kong,
 \textsuperscript{\rm $\clubsuit$}Bytedance,
 \textsuperscript{$\diamondsuit$}Stanford University,
\\
 \texttt{xut0092@link.cuhk.edu.hk}, \texttt{jksun@stanford.edu},\\
 \texttt{\{zhichao.huang, chengshanbo\}@bytedance.com, wlam@se.cuhk.edu.hk}
}
\begin{document}
\maketitle
\renewcommand{\thefootnote}{\fnsymbol{footnote}}
\footnotetext[1]{Work done when Ting Xu was interned at Bytedance.  } 
\footnotetext[2]{Corresponding authors.} 

\begin{abstract}
We present Sequential Policy Optimization for Simultaneous Machine Translation ({\methoda}), a new policy optimization framework that defines the simultaneous machine translation (SiMT) task as a sequential decision making problem, incorporating a tailored reward to enhance translation quality while reducing latency.
In contrast to popular Reinforcement Learning from Human Feedback (RLHF) methods, such as PPO and DPO, which are typically applied in single-step tasks, \methodb effectively tackles the multi-step SiMT task.
This intuitive framework allows the SiMT LLMs to simulate and refine the SiMT process using a tailored reward. We conduct experiments on six datasets from diverse domains for En $\to$ Zh and Zh $\to$ En SiMT tasks, demonstrating that {\method} consistently achieves significantly higher translation quality with lower latency. In particular, {\method} outperforms the supervised fine-tuning (SFT) model by 1.13 \footnotemark[3] points in COMET, while reducing the Average Lagging by 6.17 in the NEWSTEST2021 En $\to$ Zh dataset.  
While SiMT operates with far less context than offline translation, the SiMT results of {\method} on 7B LLM surprisingly rival the offline translation of high-performing LLMs, including Qwen-2.5-7B-Instruct and LLaMA-3-8B-Instruct.

\footnotetext[3]{\citet{kocmi-etal-2024-navigating} has indicated that an increase of 1 point in COMET represents a significant improvement.}

\end{abstract}

\section{Introduction}
\label{chap-intro}

\begin{figure*}
    \centering
    \includegraphics[width=\linewidth]{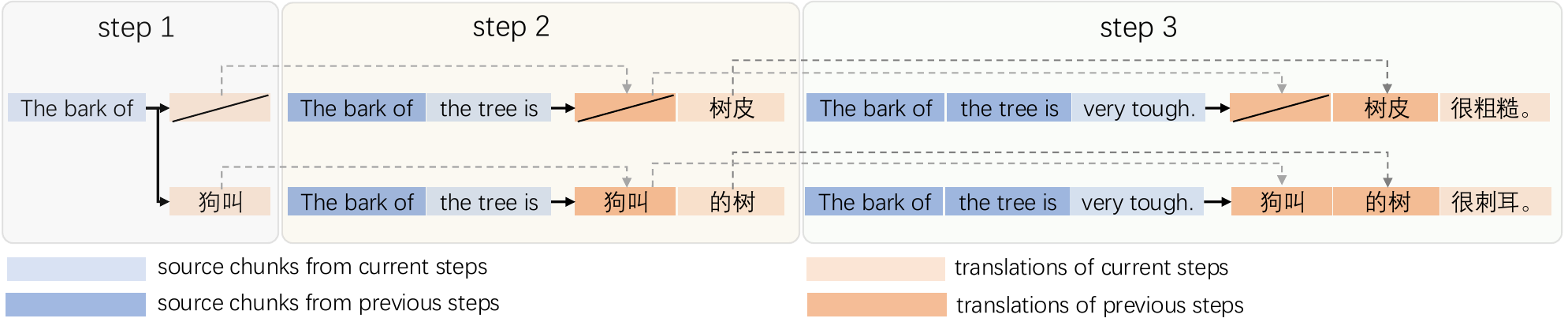}
    \caption{Two examples of SiMT, which translates streaming source texts into target texts. The source texts of SiMT are fed in a sequential manner. At each step, SiMT receives a new source text chunk and generates corresponding translations. Translations at each step can influence overall translations. 
    A forward slash (/) indicates an empty translation, which means the model chooses not to translate and waits for subsequent texts.}
    \label{fig:example}
\end{figure*}

Simultaneous Machine Translation (SiMT) has made huge progress by leveraging large language models (LLMs) \cite{cheng2024clasi, koshkin-etal-2024-transllama}. These approaches generally use partial translation data to finetune LLMs, enabling LLMs to translate partial source texts to target texts. However, the partial translation data are often generated using simple alignment tools, like heuristic methods \citep{ma-etal-2019-stacl} or attention mechanisms \citep{arivazhagan-etal-2019-monotonic}, which may introduce noise and degrade performance.

In parallel, Reinforcement Learning from Human Feedback (RLHF; \citealp{Ouyang2022TrainingLM}) has gained huge success in improving the performance of fine-tuned LLMs \citep{DeepSeekAI2025DeepSeekR1IR,yang2024qwen2,sun2025reasoning}. 
RLHF is reward-driven and does not rely on partial translation data. 
Applying these techniques to SiMT appears to be a promising approach for improving its performance. However, we find that traditional RLHF methods like PPO \citep{DBLP:journals/corr/SchulmanWDRK17} and DPO \citep{rafailov2024direct} commonly work for a \textit{single-step} process, while SiMT translates streaming inputs in a \textit{multi-step} manner. 
The multi-step dependency between the source and target in SiMT is complex. 
First, the source texts of SiMT are provided step by step, and each step's source may have ambiguous meanings that require subsequent context for clarification. For example, the first step's source (\textit{bark}) in Figure \ref{fig:example} is ambiguous, which requires subsequent source texts (\textit{tree}) to clarify. Second, previous translation results can influence the overall translation quality. For example, in the second example of Figure \ref{fig:example}, misinterpreting \textit{"the bark of"} leads to errors in the overall translations. We claim that traditional RLHF methods commonly used in single-step reveal deficiencies in modeling the complex dependence relations in the multi-step SiMT setting.

To this end, we propose a new policy optimization method, \textbf{Seq}uential \textbf{P}olicy \textbf{O}ptimization (SeqPO), and apply it to the SiMT task.
As shown in Figure \ref{fig:model}, {\methoda} defines SiMT as a sequential decision-making process. 
To simulate the SiMT process, we segment a full sentence into multiple chunks and feed these chunks to the LLM sequentially. At each step, the model evaluates the new chunk and translation history to decide whether to translate or wait for subsequent context. After completing all the steps, we construct a tailed reward to assess the entire SiMT process according to quality and latency. Finally, we optimize the SiMT process using policy gradient \citep{sutton2018reinforcement}. \methodb enables us to simulate the complex SiMT process as a multi-step decision problem and refine it based on quality and latency.

To validate the effectiveness of {\methoda}, we conduct experiments on extensive datasets for En $\to$ Zh and Zh $\to$ En SiMT tasks, including formal spoken language, informal spoken language, specialized knowledge domains, and news articles. Performance is measured using a range of comprehensive metrics.
Extensive experimental results demonstrate that {\method} not only attains superior translation quality but also reduces latency.
In low latency and high latency scenarios, the average COMET scores of {\method} are 1.3 and 1.25 points higher than those of the supervised fine-tuning (SFT) method, respectively. 
In particular, {\method} outperforms the SFT model by 1.13 points in COMET, while reducing the Average Lagging (AL) by 6.17 in NEWSTEST2021 En $\to$ Zh dataset.  
Because the SiMT task has only a limited amount of contexts while offline translation utilizes the full context, SiMT is more challenging than offline translation. Remarkably, the SiMT performance of {\method} is comparable to the offline translation performance of high-performing open-source models, such as Qwen-2.5-7B-Instruct \citep{yang2024qwen2} and LLaMA-3-8B-Instruct.
These findings underscore the effectiveness of {\methoda}.
We summarize our contribution as follows:

\noindent 1. We define the RLHF process of SiMT LLMs as a sequential decision-making process 
to model the complex dependencies among steps in SiMT.

\noindent 2. {\methoda} fuses both translation quality and latency into a reward. With a carefully designed fusion function, {\methoda} successfully improves the two metrics.

\noindent 3. Extensive experiments demonstrate the superiority of {\methoda}, which not only enhances translation quality but also reduces the latency of SiMT. Furthermore, {\methoda} achieves SiMT translation quality comparable to the offline performance of strong LLMs like Qwen-2.5-7B-Instruct.
 
\section{Sequential Policy Optimization}

\begin{figure*}[ht]
    \centering
    \includegraphics[width=\linewidth]{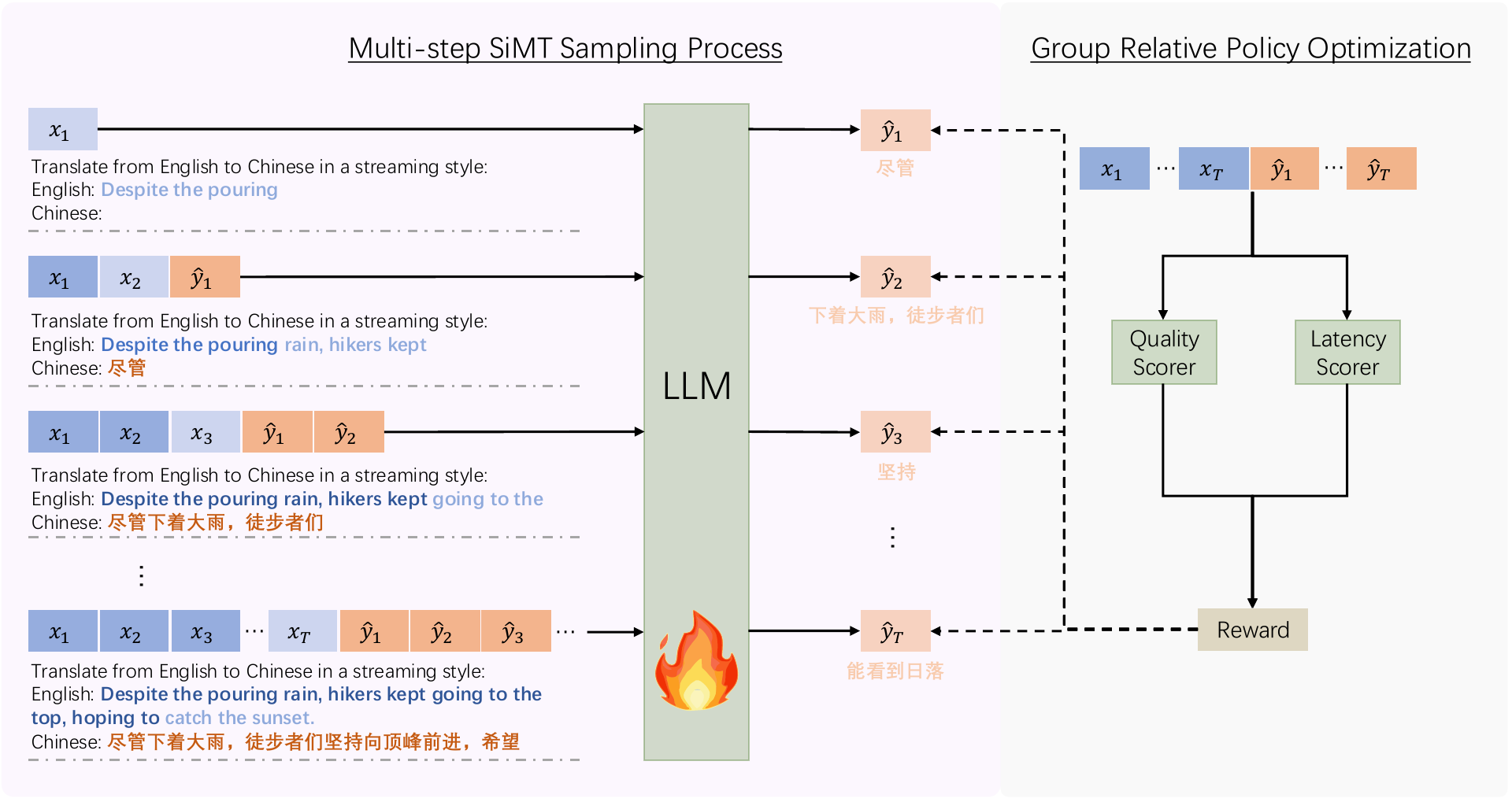} 
    \caption{Model structure of \method. We first segment a full sentence in multiple chunks. At each step, we feed a new chunk to the LLM. Concatenating new source chunk and translation history, we query the LLM to generate new translations. At the end of the translation process, we evaluate the whole translation process by quality and latency. \MyColorBox[lightblue]{Light blue} (\MyColorBox[lightorange]{Light orange}) represents the new source chunk (new translation) for each step, and \MyColorBox[mydarkblue]{dark blue} (\MyColorBox[darkorange]{dark orange}) represents the source chunks (translations) from previous steps.}
    \label{fig:model}
\end{figure*}
\begin{algorithm*}[t] 
    \caption{\method} 
    \label{alg:rlqlf} 
    \begin{algorithmic}[1]
    \REQUIRE Translation quality metrics function $f_q$, Latency metrics function $f_l$, Source sentence $\rvx$, (Optional) Reference translation $\rvy$, Initial model $\pi_\text{ref}$, Policy model $\pi_\theta$, Learning rate $\alpha$
    \ENSURE Optimized model $\pi_{\theta^*}$\\ 
    \FOR{$ \text{step} = 1$ to $N$}
        \STATE Sample $\rvx$ from training dataset and segment $\rvx$ into $T$ chunks $\rvx_1, \cdots, \rvx_T$ 
        \FOR {$t = 1$ to $T$}
            \STATE Sample $B$ translations from the policy $\pi_\theta$: 
            $\hat{y}^i_t \sim \pi_\theta(\hat{y}_t^i| x_1, \cdots, x_t; \hat{y}^i_1,, \cdots, \hat{y}^i_{t-1})$, $i=1,\cdots, B.$
        \ENDFOR
        \FOR{$i=1$ to $B$}
            \STATE Gather the SiMT process: $\hat{\rvy}^i=(\hat{y}^i_1, \cdots, \hat{y}^i_T)$
            \STATE Compute the quality and latency score: $\hat{q}^i = f_q(\rvx, \hat{\rvy}^i, \rvy)$, $\hat{L}^i = f_l(\rvx, \hat{\rvy}^i, \rvy)$
            \STATE Normalize $\hat{q}^i$, $\hat{L}^i$ to get $q^i$, $L^i$
            \STATE Calculate overall reward $r^i_T = \lambda q^i - L^i$ and  KL divergence $D^i_t = \log \frac{\pi_\theta(\hat{y}^i_t| x^i_1, \cdots, x^i_t; \hat{y}^i_1,...\hat{y}^i_{t-1})}{\pi_{\text{ref}}(\hat{y}^i_t| x^i_1, ..., x^i_t; \hat{y}^i_1,...,\hat{y}^i_{t-1})}$
        \ENDFOR
        \STATE Calculate policy gradient $\rvg = \frac{1}{B}\sum_{i=1}^B \sum_{t=1}^T (r^i_T - \beta D^i_t)\nabla_\theta \log \pi_\theta(\hat{y}^i_t| x_1, \cdots, x_t; \hat{y}^i_1,...\hat{y}^i_{t-1})$
        \STATE Update model $\theta = \theta + \alpha \rvg$
    \ENDFOR
    \end{algorithmic}
\end{algorithm*}

{\method} is a policy optimization framework that defines the SiMT task as a sequential decision making problem. It incorporates a tailored reward to enhance translation quality while reducing latency.
 This section first defines the basic components of {\method}: environment and policy. Then we describe the multi-step SiMT data sampling process. Finally, we describe how we optimize the policy model. The model architecture is illustrated in Figure \ref{fig:model}, and the algorithm is described in Algorithm \ref{alg:rlqlf}. For details on notations, please refer to Table \ref{tab:notations}.
 
\begin{table}[ht]
    \centering
    \small 
    \begin{tabular}{ll}
    \toprule
       Notation  & Meaning \\
    \midrule    
       $\mathbf{x}$  &  full source sentence \\
       $\mathbf{y}$ & full target sentence \\
       $x_t$     & source text chunk \\
       $m$  & \# words in each source text chunk \\
       $T$ & \# steps in a SiMT process \\
       $y_t$     & target text chunk \\
       $B$ & \# sampling times in each step\\
       $\pi_\theta$  & policy model \\
       $\pi_{\text{ref}}$  & reference model \\
       $f_q, f_l$ & quality and latency scorer function\\
       $\hat{q}, q$ & quality score, normalized quality score \\
       $\hat{l}, l$ & latency score, normalized latency score \\
       $r_T$ & reward \\
    \bottomrule
    \end{tabular}
    \caption{Basic notations of this work.}
    \label{tab:notations}
\end{table}

\subsection{Basic Components}
\paragraph{Environment.}
Consider a full source sentence $\mathbf{x} = (x_1, x_2, \cdots, x_T)$, where $x_t$ is a source text chunk with $m$ words, $T = \frac{|\mathbf{x}|}{m}$. At each time step, the environment emits a new text chunk in the full source sentence. After the environment emits the last text chunk, it reaches a terminal state and the SiMT process ends.

\paragraph{Policy.} Following \citet{cheng2024clasi} and \citet{koshkin-etal-2024-transllama}, we employ an LLM as the policy model $\pi_\theta$ to generate translations. At time step $t$, the input to the LLM is based on existing source text chunks $x_{1:t} = (x_1, x_2, \cdots, x_t)$ and previous translation history $\hat{y}_{1:t-1}=(\hat{y}_1, \hat{y}_2, \cdots, \hat{y}_{t-1})$. Concatenating all existing source text chunks and previous translations, the policy model produces translations as follows:
\begin{equation}
    \hat{y}_t \sim \pi_\theta(\hat{y}_t|x_1, \cdots, x_t, \hat{y}_1,\cdots, \hat{y}_{t-1}).
    \label{eq:policy}
\end{equation}
Although every source chunk $x_t$ is of pre-defined length $m$, the length of $\hat{y}_t$ is totally decided by the policy model. If the policy model decides not to translate and waits for more context, $\hat{y}_t$ will be empty, meaning its length is $0$. If the policy model chooses to translate, $\hat{y}_t$ will consist of all the tokens generated by the policy model. 
This contrasts with rule-based methods. We allow the policy model to determine when to start translating and how much content to translate based on the context. This makes our policy more flexible than previous methods.
 
\subsection{Multi-step SiMT Sampling Process}
Traditional methods often fine-tune LLMs using partial translation data, which are derived from aligning full translation pairs. However, the alignment tools are often simple, such as  heuristic methods or attention mechanisms, which may introduce noise and degrade performance.
Instead of aligning the source and target sentences, we directly simulate and refine the SiMT process. We follow the multi-step nature of SiMT and conduct a multi-step SiMT sampling process as follows: 
\begin{itemize}
    \item The environment converts the source sentence $\mathbf{x}$ into $T$ chunks: $\mathbf{x} = (x_1, x_2, \cdots, x_T)$, where $T$ is the number of chunks. For example, we convert the source sentence in Figure \ref{fig:model} into five chunks: \textit{("Despite the pouring", "rain, hikers kept", "going to the", "top, hoping to", "catch the sunset")}.
    \item Initially, the environment emits the first text chunk $x_1$, and the policy produces $B$ candidate translations $\hat{y}_1^i$ based on $x_1$: $\hat{y}_1^i \sim \pi_\theta(x_1)$, for $i=1,\cdots, B$, where $B$ is the number of sampling times. As shown in Figure \ref{fig:model}, the model translates the first chunk \textit{"Despite the pouring"} into \textit{"{\small\begin{CJK}{UTF8}{gbsn}尽管\end{CJK}}"}.

    \item At each subsequent time step t, the environment emits a new text chunk, $x_t$. The policy model then produces translations based on Equation (\ref{eq:policy}), considering both the new text chunk and the translation history. For example, as illustrated in step 2 of Figure \ref{fig:model}, concatenating the previous text chunk ("\textit{Despite the pouring}") with the current text chunk ("\textit{rain hikers kept}") yields the source text ("\textit{Despite the pouring rain, hikers kept}"). This is then concatenated with the previous translation ("{\small\begin{CJK}{UTF8}{gbsn}\textit{尽管}\end{CJK}}"). We fill in the source texts and translation into the template, construct a prompt for the model to generate translation, and obtain the result ({\small\begin{CJK}{UTF8}{gbsn}\textit{"下着大雨，徒步者们"}\end{CJK}}). 

    \item Keep running the last step until all source texts are fully translated, i.e., $t > T$. 
    
\end{itemize}

In the end, by aggregating all the translation steps, we have $\hat{\mathbf{y}}^i = (\hat{y}_1^i, \hat{y}_2^i, \cdots, \hat{y}_T^i)$. For example in Figure \ref{fig:model}, the final translation is {\small\begin{CJK}{UTF8}{gbsn}\textit{"尽管下着大雨，徒步者们坚持向顶峰前进，希望能看到日落"}\end{CJK}}.

\subsection{Reward}
After the multi-step sampling process, we evaluate the entire SiMT process and further refine it. Concretly, an accurate final reward is provided to policy model $\pi_\theta$ at the last step $T$ as $r_T$, and the policy model can adopt $r_T$ to optimize its sequential decision from step $1$ to step $T$. Specifically, we assess SiMT's performance from two perspectives: quality and latency.
Translation quality can be evaluated using existing metrics for machine translation, such as COMET \citep{rei2020comet}, BLEURT \citep{sellam-etal-2020-bleurt}, or the RLHF reward model \citep{Ouyang2022TrainingLM}. Latency can be measured using metrics like Average Lagging (AL; \citealp{ma-etal-2019-stacl}) and Length-Adaptive Average Lagging (LAAL; \citealp{papi-etal-2022-generation}). 

Our primary objective is to identify a translation process that achieves not only high quality but also maintains low latency. 
However, there is an inevitable trade-off between translation quality and latency. On one hand, a conservative policy that waits longer may have higher translation quality. On the other hand, a radical policy with small latency may be short of translation quality. 
In order to balance these two metrics and unify them into the same scale, we propose to incorporate reward normalization and latency truncation into a fused reward to measure both quality and latency. Specifically, we define the reward as follows: 
\begin{equation}
\begin{aligned}
    \hat{q}^i &= f_q(\mathbf{x}^i, \hat{\mathbf{y}}^i, \mathbf{y}), \quad \hat{L}^i = f_l(\mathbf{x}^i, \hat{\mathbf{y}}^i, \mathbf{y}), \\
    q^i &= \frac{\hat{q}^i - \text{mean}(\{\hat{q}^1,\cdots, \hat{q}^B\})}{\text{std}(\{\hat{q}^1,\cdots, \hat{q}^B\})},\\
    L^i &= \max(m, \frac{\hat{L}^i - \text{mean}(\{\hat{L}^1, \cdots, \hat{L}^B\})}{\text{std}(\{\hat{L}^1, \cdots, \hat{L}^B\})}), \\
    r^i_T &= \lambda q^i - L^i,
\end{aligned}
\end{equation}
where $f_q$ and $f_l$ are quality and latency scorer functions, $\mathbf{y}$ is the gold translation, $\lambda$ is a hyper-parameter that decides the trade-off between quality and latency, the superscript $i$ means the $i$-th sample from the mini-batch. In the above equation, we normalize the values of quality and latency because they have different scales. By nomalization, we can convert these two metrics to the same scale, thereby facilitating their fusion. And we truncate the $L$ by the chunk size of $\mathbf{x}$ to avoid overfitting to the latency score. In addition to the reward, we also add commonly used KL constraint \citep{DBLP:journals/corr/SchulmanWDRK17} to keep the policy model stable in the training process. Thus, the final objective function of {\methoda} is 
\begin{equation}
    \small
    \begin{aligned}
    J(\pi_\theta) = &\mathbb{E}_{\hat{y}_1,\cdots, \hat{y}_T \sim \pi_\theta(\cdot), \mathbf{x} \sim p_\text{data}}  \sum_{t=1}^T [r_T \\
    &- \beta \log \frac{\pi_\theta(\hat{y}_t| x_1,\cdots, x_t;\hat{y}_1, \cdots, \hat{y}_{t-1} )}{\pi_{\text{ref}}(\hat{y}_t| x_1,\cdots, x_t; \hat{y}_1, \cdots, \hat{y}_{t-1})}],
\end{aligned}
\end{equation}
where $p_\text{data}$ is the training dataset.

\subsection{Optimization}
As for the optimization method, our approach can be applied to various policy gradient optimization methods. In this paper, 
Group Relative Policy Optimization (GRPO; \citealp{shao2024deepseekmath}), which has demonstrated its effectiveness and efficiency in various large language models, such as DeepSeekMath \citep{shao2024deepseekmath} and DeepSeek-R1 \citep{DeepSeekAI2025DeepSeekR1IR}, is selected as our optimization method. 
Specifically, we sample $B$ trajectories for each to calculate the baseline reward. In {\methoda}, we choose GRPO over the popular PPO \citep{DBLP:journals/corr/SchulmanWDRK17}  because of the following reasons:

\noindent 1. \textbf{Resources Efficiency}: GRPO utilizes a grouped average as its baseline, whereas PPO incorporates a new critic model for baseline computation, which increases memory and computational overhead. In our context, translation quality and latency require two separate critic models, which makes the memory requirements unaffordable. 

\noindent 2. \textbf{Accurate Metric}: In SiMT, latency is a rule-based metric. PPO uses a neural reward model may introduce more noise and complicate the training pipeline \citep{DeepSeekAI2025DeepSeekR1IR}.

\begin{figure*}[ht]
    \centering
    \begin{subfigure}[t]{0.32\textwidth}
        \centering
        \includegraphics[width=\linewidth]{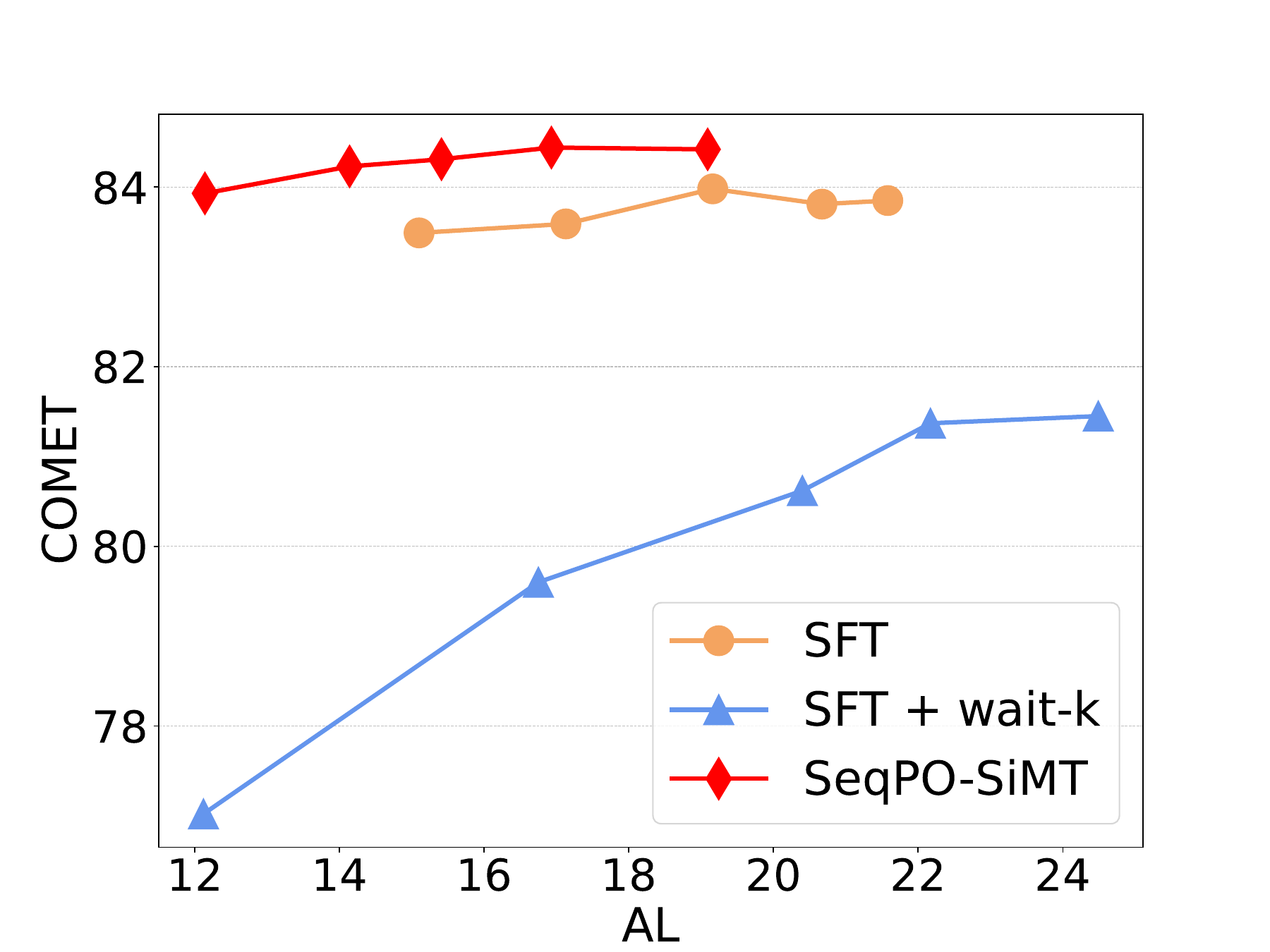}
        \caption{REALSI Zh $\to$ En}
        \label{fig:clasi-zh2en}
    \end{subfigure}
    \hfill
    \begin{subfigure}[t]{0.32\textwidth}
        \centering
        \includegraphics[width=\linewidth]{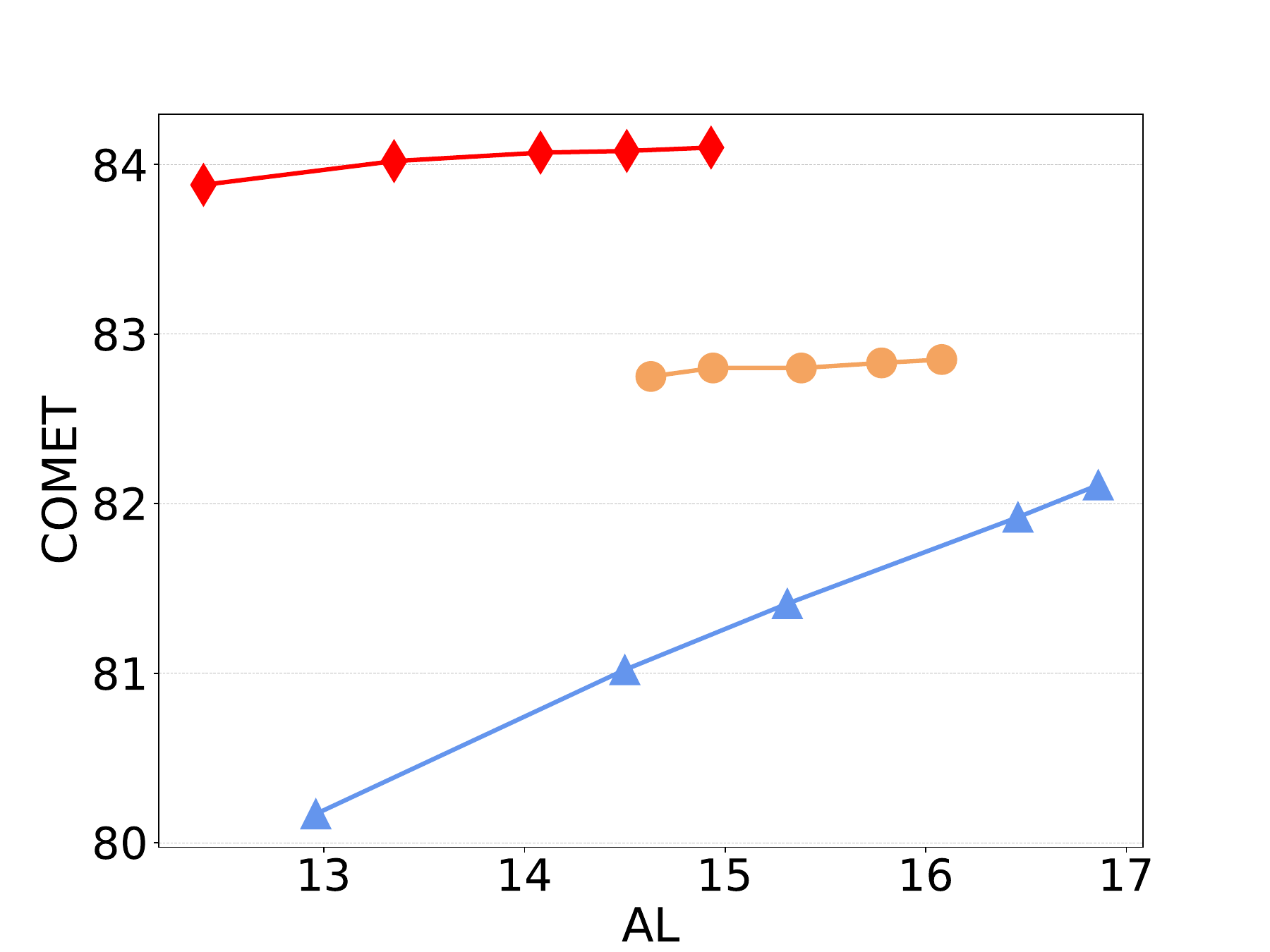}
        \caption{COVOST Zh $\to$ En}
        \label{fig:covost-zh2en}
    \end{subfigure}
    \hfill 
    \begin{subfigure}[t]{0.32\textwidth}
        \centering
        \includegraphics[width=\linewidth]{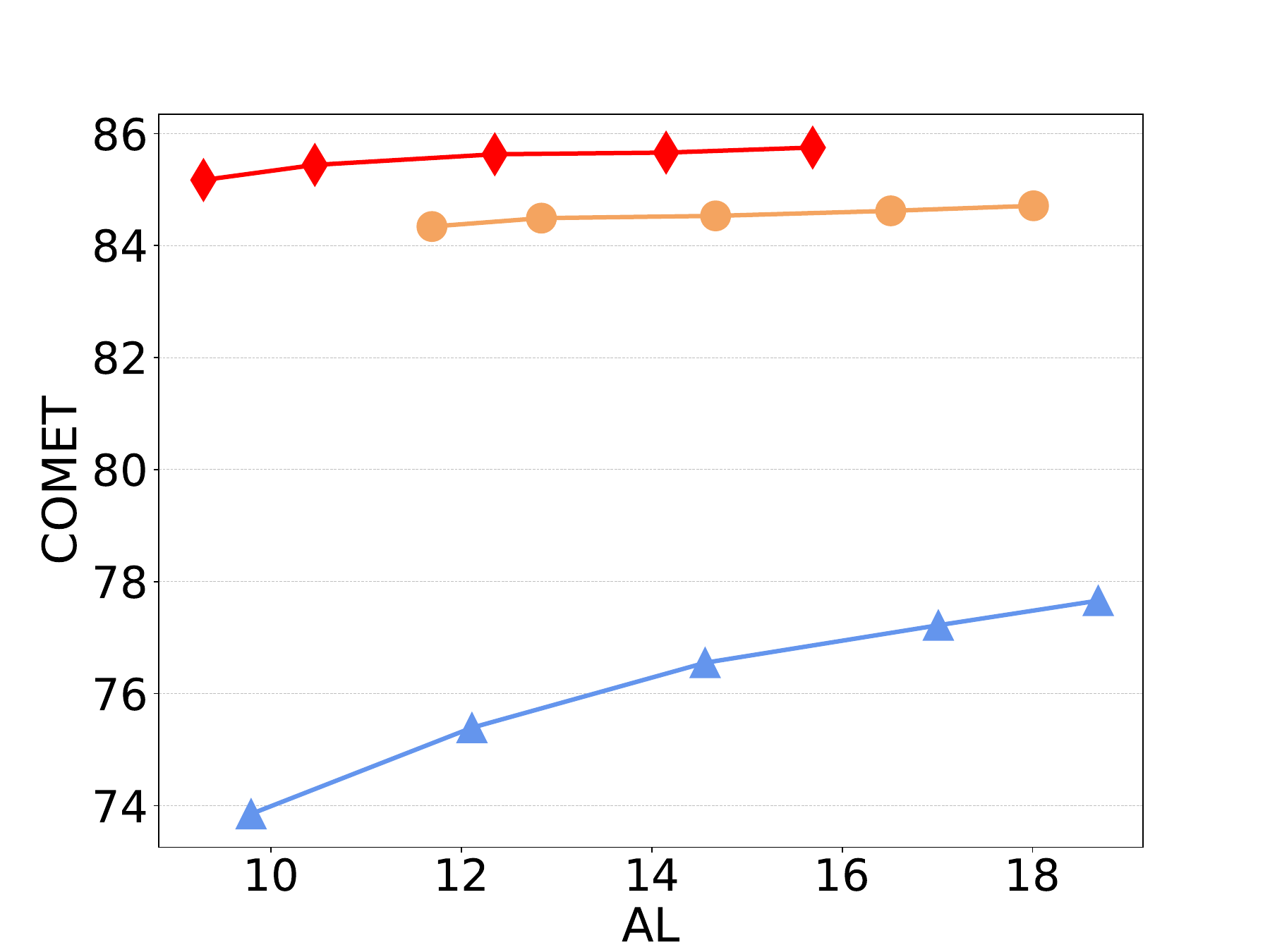}
        \caption{NEWSTEST2021 Zh $\to$ En}
        \label{fig:newstest-zh2en}
    \end{subfigure}

    \begin{subfigure}[t]{0.32\textwidth}
        \centering
        \includegraphics[width=\linewidth]{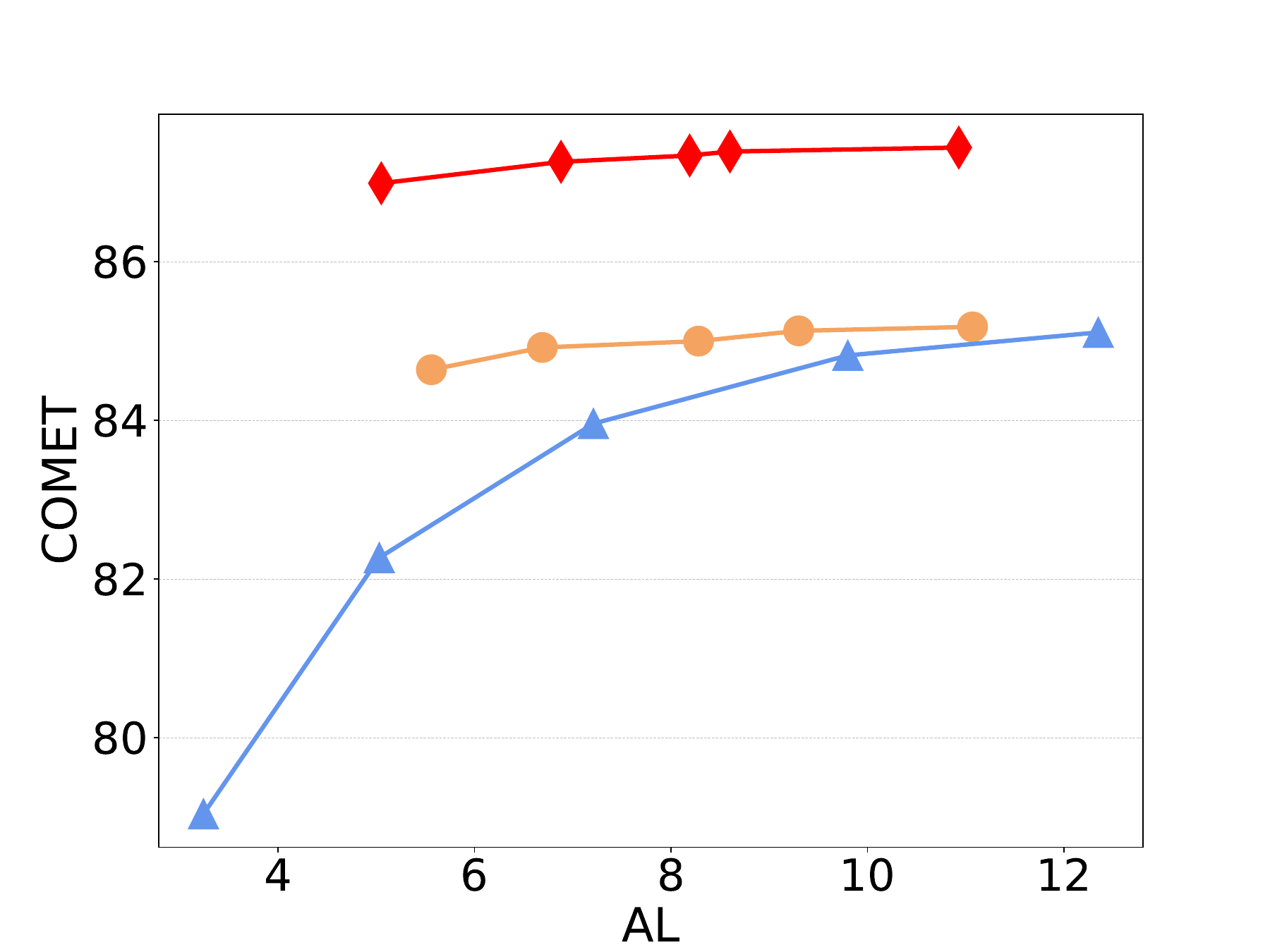}
        \caption{REALSI En $\to$ Zh}
        \label{fig:clasi-en2zh}
    \end{subfigure}
    \hfill
    \begin{subfigure}[t]{0.32\textwidth}
        \centering
        \includegraphics[width=\linewidth]{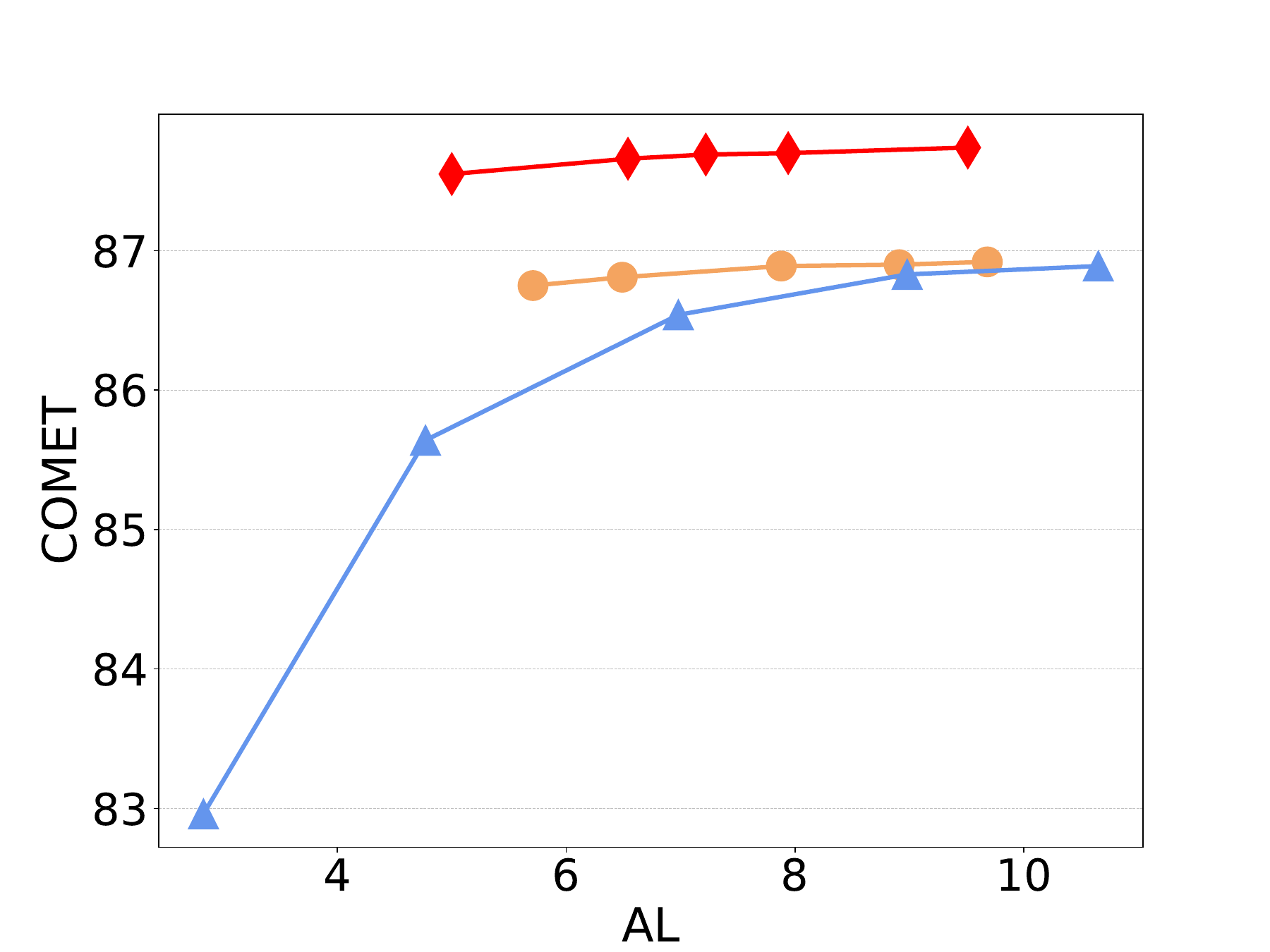}
        \caption{MUSTC En $\to$ Zh}
        \label{fig:mustc-en2zh}
    \end{subfigure}
    \hfill
    \begin{subfigure}[t]{0.32\textwidth}
        \centering
        \includegraphics[width=\linewidth]{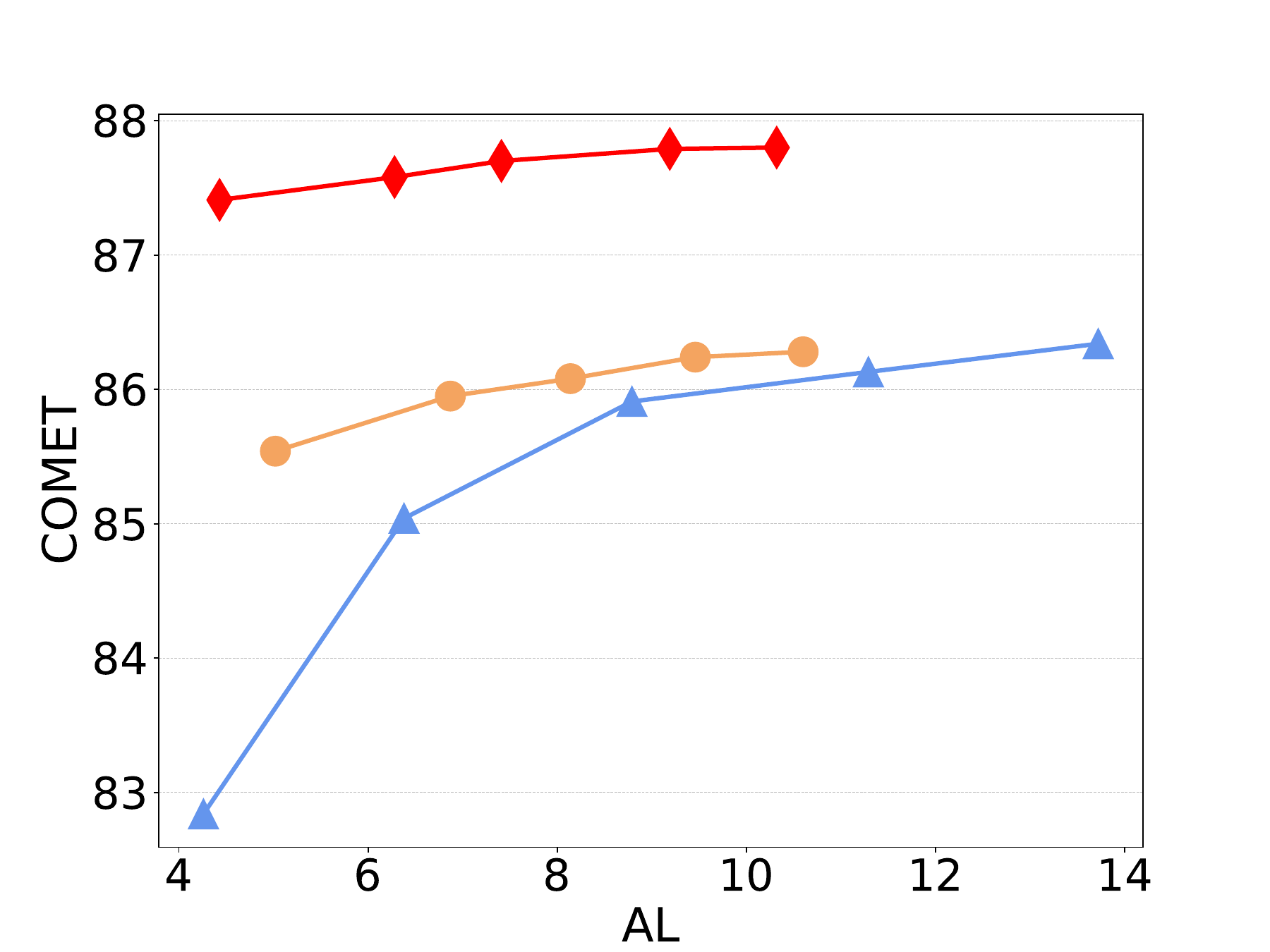}
        \caption{NEWSTEST2021 En $\to$ Zh}
        \label{fig:newstest-en2zh}
    \end{subfigure}
    \caption{COMET v.s. AL on Zh $\to$ En and En $\to$ Zh SiMT tasks. }
    \label{fig:main-comet-al}
\end{figure*}

\begin{table*}[ht]
    \centering
    \small
    \setlength\tabcolsep{2.0pt}

    \begin{tabular}{llcccccccccc}
        \toprule
        \multirow{2}{*}{Dataset}& \multirow{2}{*}{Method} & \multicolumn{5}{c}{Low latency} & \multicolumn{5}{c}{High latency}\\
        \cmidrule(lr){3-7}\cmidrule(lr){8-12}
        & &  BLEURT $\uparrow$ & COMET$\uparrow$ & GPT-4 $\uparrow$ & AL $\downarrow$ & LAAL $\downarrow$ & BLEURT $\uparrow$ & COMET $\uparrow$ & GPT-4 $\uparrow$ & AL $\downarrow$ & LAAL $\downarrow$ \\
        \midrule 
        \multirow{3}{*}{REALSI} & SFT & 64.14 &	83.49 & 83.24 &	15.1 &	15.87 & 64.8	& 83.77	& 84.07 & \textbf{18.27} &	\textbf{18.94}\\
        & SFT+wait-$k$ & 59.37 & 79.6 & 78.9 &  16.75 & 16.97 & 61.2 & 80.97 & 79.87 & 	22.17& 22.37\\
        & \method & \textbf{65.93}	& \textbf{84.23} & \textbf{85.49} &	\textbf{14.14}&\textbf{14.59} & \textbf{66.24} & 	\textbf{84.42} & \textbf{85.92} & 19.09&19.44\\
        \midrule
        \multirow{3}{*}{COVOST} & SFT & 60.17	& 82.75	& 75.47 & 14.63	& 14.72 &60.33	& 82.85	& 75.86  & 16.08	& 16.13\\
        & SFT+wait-$k$ & 57.06	&80.17&71.47 &	12.96&	13.06 & 59.16	&81.92	&73.93  &16.46	&16.49\\
        & \method &\textbf{63.01}	& \textbf{83.95} &	\textbf{79.86} & \textbf{12.91} &	\textbf{13.01} & \textbf{63.28}	& \textbf{84.1} & \textbf{80.13} &\textbf{14.93}& \textbf{14.99}\\
        \midrule
        \multirow{3}{*}{NEWS} &SFT & 65.01	&84.34 &86.02&	10.18 & 	12.94 & 65.69& 84.7 &86.54&	17.18&18.33\\
        & SFT+wait-$k$ & 50.53	& 73.85	&72.99& 9.79& 9.93 &55.03 &	77.66&78.62 & 18.69 & 18.91\\
        & \method & \textbf{66.67}	& \textbf{85.17} &  \textbf{87.67}&	\textbf{9.29} &\textbf{9.92} & \textbf{67.94}	&\textbf{85.75} & \textbf{89.17}	&\textbf{15.69}	&\textbf{16.34}\\
        \bottomrule
    \end{tabular}
    \caption{Detailed results in low and high latency levels on Zh $\to$ En SiMT tasks. NEWS is an abbreviation for NEWSTEST2021. The best results are highlighted in bold.}
    \label{tab:main-results-zh2en}
\end{table*}

\begin{table*}[ht]
    \centering
    \small
    \setlength\tabcolsep{2.0pt}
    \begin{tabular}{llcccccccccc}
        \toprule
        \multirow{2}{*}{Dataset}& \multirow{2}{*}{Method} & \multicolumn{5}{c}{Low latency} & \multicolumn{5}{c}{High latency}\\
        \cmidrule(lr){3-7}\cmidrule(lr){8-12}
        & &  BLEURT $\uparrow$ & COMET$\uparrow$ & GPT-4 $\uparrow$ & AL $\downarrow$ & LAAL $\downarrow$ & BLEURT $\uparrow$ & COMET $\uparrow$ & GPT-4 $\uparrow$ & AL $\downarrow$ & LAAL $\downarrow$ \\
        \midrule 
        \multirow{4}{*}{REALSI} & SFT & 61.84	& 84.64	& 87.45 &  5.56	&5.77& 63.19	& 85.18	&88.11 & 10.34	& 11.22\\
        & SFT + wait-$k$ & 58.92	 & 82.27 &  84.06 & \textbf{5.03}	& \textbf{5.14} & 62.46	& 84.82 & 88.33	& \textbf{9.8}	& \textbf{9.96}\\
        & \method & \textbf{64.84}	 & \textbf{86.99} & \textbf{87.53} &5.05	&\textbf{5.14} & \textbf{66.41} &  \textbf{87.44} & \textbf{89.07}&10.93 & 11.04\\
        \midrule
        \multirow{4}{*}{MUSTC} &SFT &65.84 & 	86.75& 91.84 & 5.71	& 5.95 & 66.12 &	86.92 &92.45 &	9.68 &	9.84\\
        & SFT + wait-$k$ & 63.94	&85.64	& 89.15&\textbf{4.77} & \textbf{4.95} & 66.04 & 	86.83 &92.24& 	\textbf{8.98}& \textbf{9.16}\\
        & \method & \textbf{66.76}	&\textbf{87.55}& \textbf{92.1} & 	5	&5.15 & \textbf{67.46}	&\textbf{87.72} & \textbf{93.18} &	9.51 & 9.62\\
        \midrule
        \multirow{4}{*}{NEWS} & SFT & 61.12 &	85.54 & 90.99& 	5.02	& 5.9 & 62.28 &	86.28 & 92.28 & 10.6 &11.29\\
        & SFT + wait-$k$ & 57.1	& 82.84	&83.58& \textbf{4.26}& \textbf{4.85} & 62.25	&86.13 &92.2&11.29	&11.9\\
        & \method & \textbf{63.37}	& \textbf{87.41} &\textbf{91.63}	& 4.43& 5.02 & \textbf{64.62} &	\textbf{87.78} & \textbf{93.31}& \textbf{10.32} & \textbf{10.83}\\
        \bottomrule
    \end{tabular}
    \caption{Detailed results in different latency level on En $\to$ Zh SiMT tasks. NEWS is an abbreviation for NEWSTEST2021. The best results are highlighted in bold.}
    \label{tab:main-results-en2zh}
\end{table*}
\section{Experimental Setup}

\paragraph{Models and Training Data.}
We focus on Chinese and English, a language pair with significant structural differences. 
For the Zh $\to$ En setup, we utilize WeNet \citep{zhang2022wenet} and WMT19 \citep{barrault-etal-2019-findings} as our training data. 
For the En $\to$ Zh setup, we employ GigaST \citep{chen2021giga, gigast} as the training data. 
We use Qwen-2.5-7B \citep{yang2024qwen2} as our backbone.

Because the base LLM lacks SiMT capabilities, we initially construct a dataset for warm-up, the process is similar to the previous works \citep{koshkin-etal-2024-transllama, cheng2024clasi}. 
Specifically, we randomly select 40,000 data samples from the training datasets and construct partial translation data by prompting LLMs. Details about the SFT data construction process is shown in Appendix \ref{appendix:sft-data}.

\paragraph{Evaluation Benchmarks.} 
To comprehensively validate the effectiveness of \method, we conduct experiments on datasets from diverse domains.  For the Zh $\to$ En setup, we evaluate \methodb on COVOST \citep{wang-etal-2020-covost}, REALSI  \citep{cheng2024clasi}, and NEWSTEST2021 on Zh $\to$ En, including specialized knowledge, informal spoken language, and news articles. For the En $\to$ Zh setup, we evaluate \methodb on REALSI, MUSTC \citep{di-gangi-etal-2019-must}, and NEWSTEST2021, including informal spoken language, formal spoken language, and news articles. 

\paragraph{Implementation Details.} 
We set the hyperparameter as follows: For $\lambda$ parameter, we sample 50 SiMT data and score them with a range of $\lambda$ values, then we manually evaluate the scoring performance for each $\lambda$ value and select $\lambda=2$ because it can balance between quality and latency. For other hyper-parameters, we set $B=5$, $\beta=0.02$ for En $\to$ Zh, $\beta=0.1$ for Zh $\to$ En. COMET and AL are chosen as the translation quality reward and latency reward, respective. Due to space limitations, we put other implementation details in Appendix \ref{appendix:exp-setup}.

\paragraph{Evaluation Metrics.}
We measure the performance through comprehensive metrics of translation quality and latency. For translation quality, we employ COMET \citep{rei2020comet}, BLEURT \citep{sellam-etal-2020-bleurt}, and GPT-4 as metrics. Detailed prompt for GPT-4 is described in Appendix \ref{appendix:exp-setup}. For latency, we use Average Lagging (AL; \citealp{ma-etal-2019-stacl}) and Length-Adaptive Average Lagging (LAAL; \citealp{papi-etal-2022-generation}) as metrics.

\paragraph{Baselines.} 
To demonstrate the effectiveness of our method, we compare the results with the SFT method and wait-$k$ \citep{ma-etal-2019-stacl}, a commonly used method in SiMT.
\begin{itemize}
    \item \textbf{SFT} trains exclusively on partial translation SFT data, which is the same as the SFT data of \method.
    \item \textbf{SFT + wait-$k$} uses the same model as SFT. During inference, it uses the wait-k policy \citep{ma-etal-2019-stacl}. When receiving the first $k$ tokens, it waits and does not generate any tokens. After the first $k$ tokens, every time it receives a new token, it will produce a new token for translation.
\end{itemize}

\section{Experimental Results}
In this section, we present the main results of our experiments, emphasizing the consistent and significant superior performance of {\method} across various benchmarks and metrics. We compare with offline translations to emphasize the high translation quality of \method. Then we provide an in-depth understanding of how {\method} concurrently improves quality and latency. 

\subsection{Main Results}
\paragraph{\methodb consistently and significantly outperforms other methods in both quality and latency.}
The COMET and AL performance for different methods are demonstrated in Figure \ref{fig:main-comet-al}. The results show that {\methodb} consistently achieves a superior translation quality across all latency levels and all datasets, particularly in the low latency level. Other figures about COMET v.s. LAAL, BLEURT v.s. AL, and BLEURT v.s. LAAL are available in appendix \ref{appendix:main-results}.

We provide detailed numerical results in Table \ref{tab:main-results-zh2en} and Table \ref{tab:main-results-en2zh}. Specifically, we first divide the latency into two groups, low latency and high latency. To avoid overfitting the model to a specific metric, we provide many metrics for quality and latency. In both high-latency and low-latency scenarios, the translation quality of \methodb is significantly higher than that of other methods, achieving consistent improvements in BLEURT, COMET, and GPT-4. On average, The COMET scores of {\method} are 1.3 and 1.25 higher than those of the supervised finetuning (SFT) method in low latency and high latency scenarios, respectively. \citet{kocmi-etal-2024-error} shows that an increase of 1 point in COMET represents a significant improvement, so the progress we made is impressive.
In particular, {\method} outperforms the SFT model by 1.13 points in COMET, while decreasing the AL by 6.17 in NEWSTEST2021 En $\to$ Zh. {\methoda} also achieves superior performance on BLEURT and GPT-4's evaluation, indicating that {\methoda} genuinely improves translation quality rather than overfitting to the COMET metric.

\paragraph{\methodb achieves more stable COMET with varied latency.}
As shown in Figure \ref{fig:main-comet-al}, as the latency increases, {\method} exhibits more stable performance compared to the wait-$k$ strategy in terms of COMET scores. Particularly in low-latency scenarios, the translation quality of our method is significantly higher than that of the wait-$k$ strategy.

\subsection{Comparison with Offline Models}
\paragraph{\methodb outperforms offline SFT model and LLaMA-3-8B-Instruct, achieving comparable results to the offline Qwen-2.5-7B-Instruct.}

As most of the previous SiMT algorithms \citep{koshkin-etal-2024-transllama, yu2025simulpl, yu-etal-2024-self, guo-etal-2024-decoder} use different benchmarks with different base models, it is hard to fairly compare previous methods in our setting. To further show that {\methoda} achieves SoTA performance in 7B LLMs, we compare the SiMT results of {\methodb} with the offline results of the high-performing open-source model, i.e., LLaMA3-8B-Instruct and Qwen2.5-7B-Instruct. For the SiMT translation results, we use results with low latency.
The results are shown in Table \ref{tab:cmp-offline}. We can see that:
{\methodb} even surpasses the translation quality of offline SFT model, Qwen2.5-7B-Instruct and LLaMa3-8B-Instruct, indicating that our method significantly boosts translation quality and achieving the SoTA performance in 7B model size.

\begin{table}[ht]
    \centering
    \small
    \begin{tabular}{lccccc}
      \toprule
      & BLEURT & COMET & GPT-4 \\
      \midrule
      
      & \multicolumn{3}{c}{MUSTC En $\to$ Zh}\\
      SFT & 65.84 & 86.75 & 92.27\\
      offline SFT & \underline{66.28}	& \underline{86.94}&	\underline{92.61}\\
      offline LLaMa3 & 60.43 & 83.78 & 86.98\\
      offline Qwen2.5 & 65.47	&86.49	&91.97\\
      \method & \textbf{66.76} & \textbf{87.55} & \textbf{92.7}\\
      \midrule
       & \multicolumn{3}{c}{REALSI Zh $\to$ En}\\
      SFT & 64.14 & 83.49 & 83.74\\
      offline SFT  & \underline{65.06}	&\underline{83.92}	&83.72\\
      offline LLaMa3 &  63.27 & 81.19 & 85.66\\
      offline Qwen2.5 & 64.08	 & 82.14	& \textbf{85.79}\\
      \method & \textbf{65.93} & \textbf{84.23} & \underline{85.59}\\
      \bottomrule
    \end{tabular}
    \caption{Comparison of translation quality between {\method}'s SiMT results and other LLMs' offline translation results. The best results and the second-best are highlighted by bold and underscore respectively.}
    \label{tab:cmp-offline}
\end{table}

\subsection{In-depth Analysis of Quality and Latency}
\begin{figure}[ht]
    \centering
    \begin{subfigure}[t]{0.23\textwidth}
        \centering
        \includegraphics[width=\linewidth]{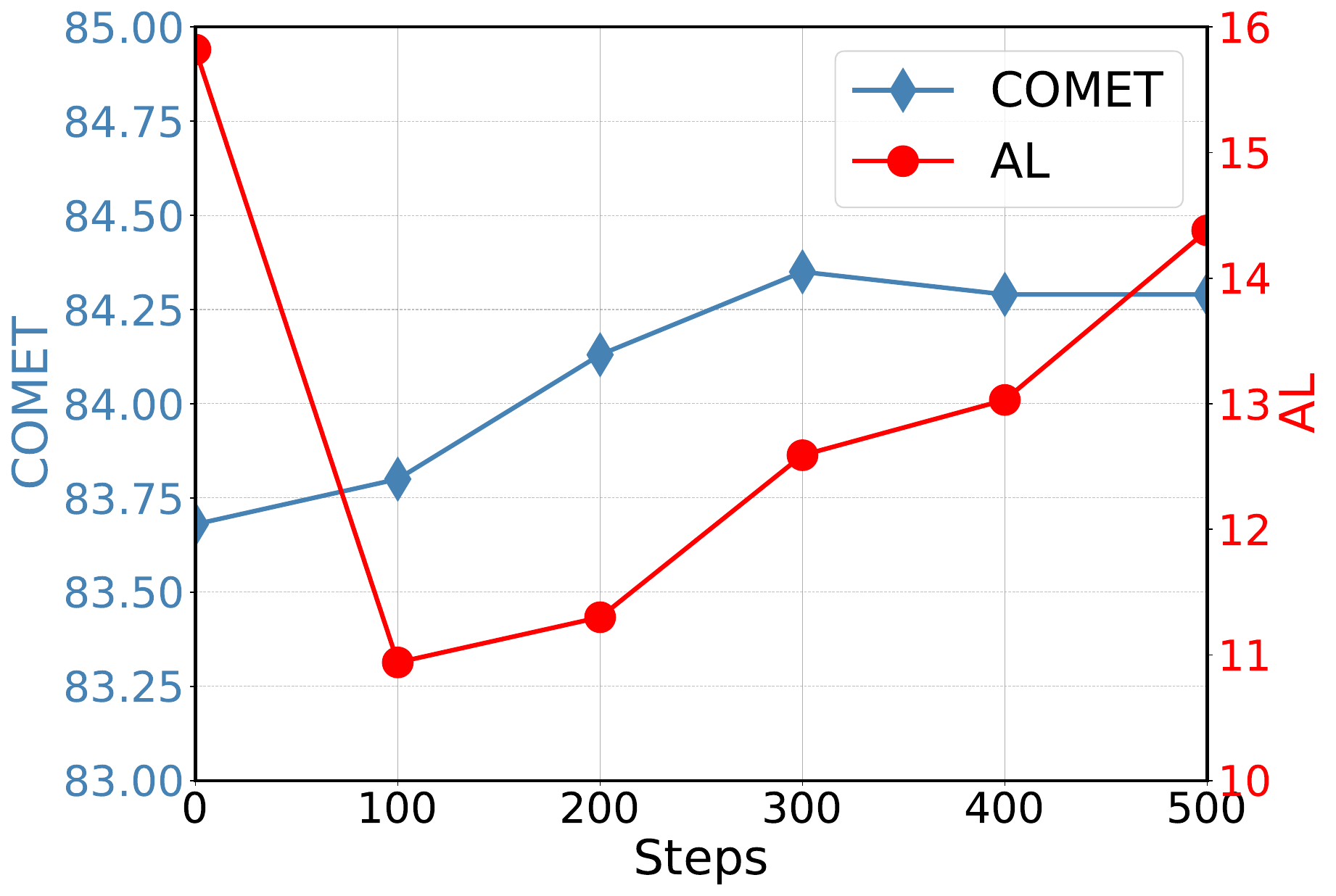}
        \caption{Change of COMET and AL during Training.}
        \label{fig:comet-al-train}
    \end{subfigure}
    \hfill
    \begin{subfigure}[t]{0.23\textwidth}
        \centering
        \includegraphics[width=\linewidth]{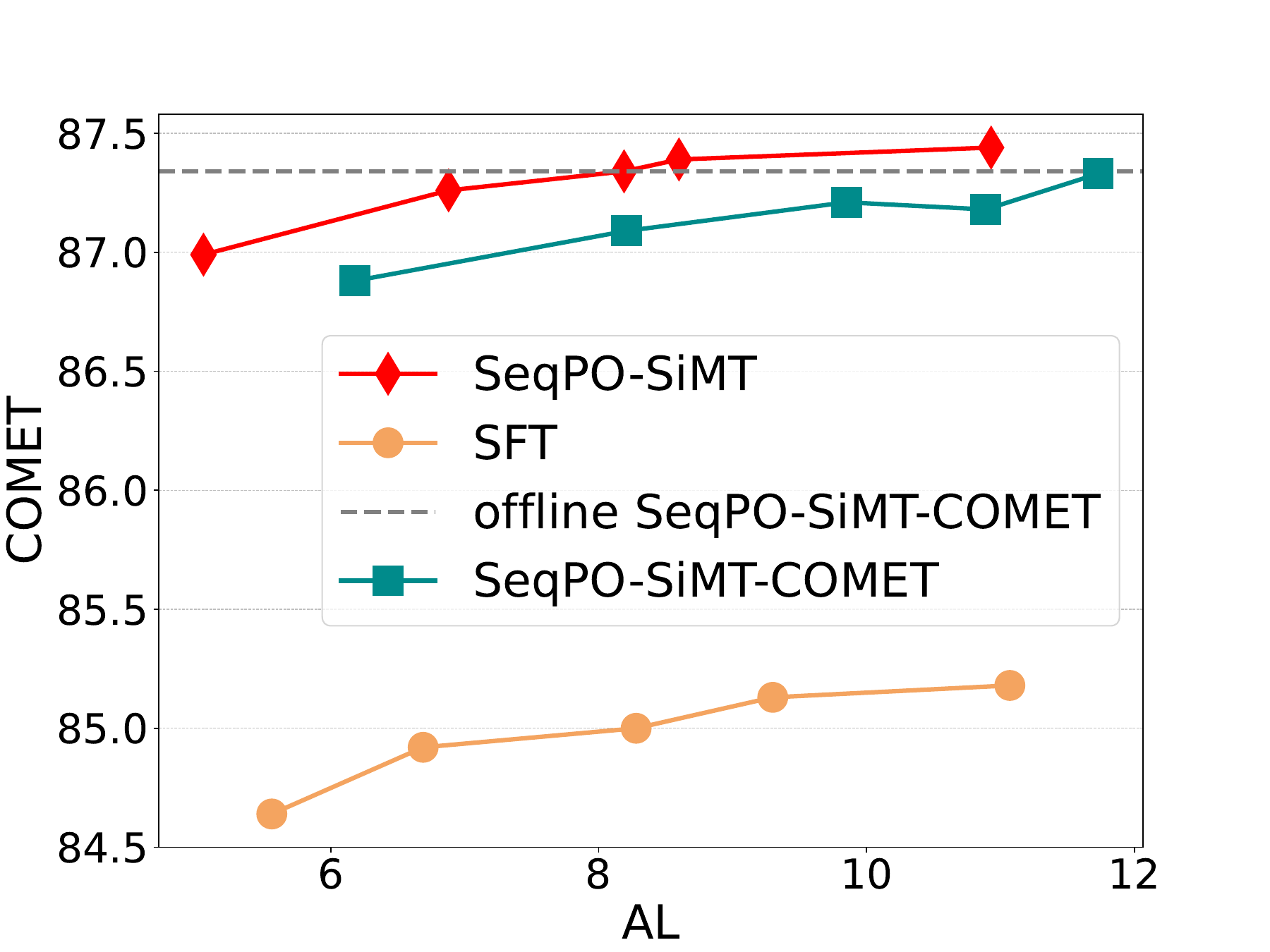}
        \caption{Comparison with only optimizing COMET.}
        \label{fig:ablation}
    \end{subfigure}
    \caption{In-depth analysis of quality and latency.}
\end{figure}

\paragraph{Change of COMET and AL during training.} We further study the changes in COMET and AL during the training process, as shown in Figure \ref{fig:comet-al-train}. During training, AL first decreases rapidly and then slowly increases, while COMET initially rises and then stabilizes. We believe this is because AL is easy to fit (if the model outputs many meaningless tokens each time it receives input, AL will also decrease). Therefore, during the training process, the model first reduces latency. Then to increase the overall reward, the model further refines its translation quality.

\paragraph{{\method} boosts both quality and latency without compromise.} We explore the maximum translation quality achievable when we only optimize translation quality, i.e., only optimize the COMET score (\method-COMET). The results, as shown in Figure \ref{fig:ablation}, indicate that {\method} has lower latency than \method-COMET at the same translation quality. Notably, different from previous methods \citep{koshkin-etal-2024-transllama, zhang-2024-guided} which sacrifice quality or latency, {\method} can achieve comparable translation quality with offline \method-COMET, demonstrating that our method effectively boosts quality and latency without compromise.

\subsection{Ablation Study on Reward}
\label{appendix:ablation}

\begin{figure}[ht]
    \centering
    \begin{subfigure}[t]{0.23\textwidth}
        \centering
        \includegraphics[width=\linewidth]{figures/Clasi_zh2en_comet_al.pdf}
        \caption{Change of COMET and AL with truncation module during Training.}
        \label{fig:wotruncation}
    \end{subfigure}
    \hfill
    \begin{subfigure}[t]{0.23\textwidth}
        \centering
        \includegraphics[width=\linewidth]{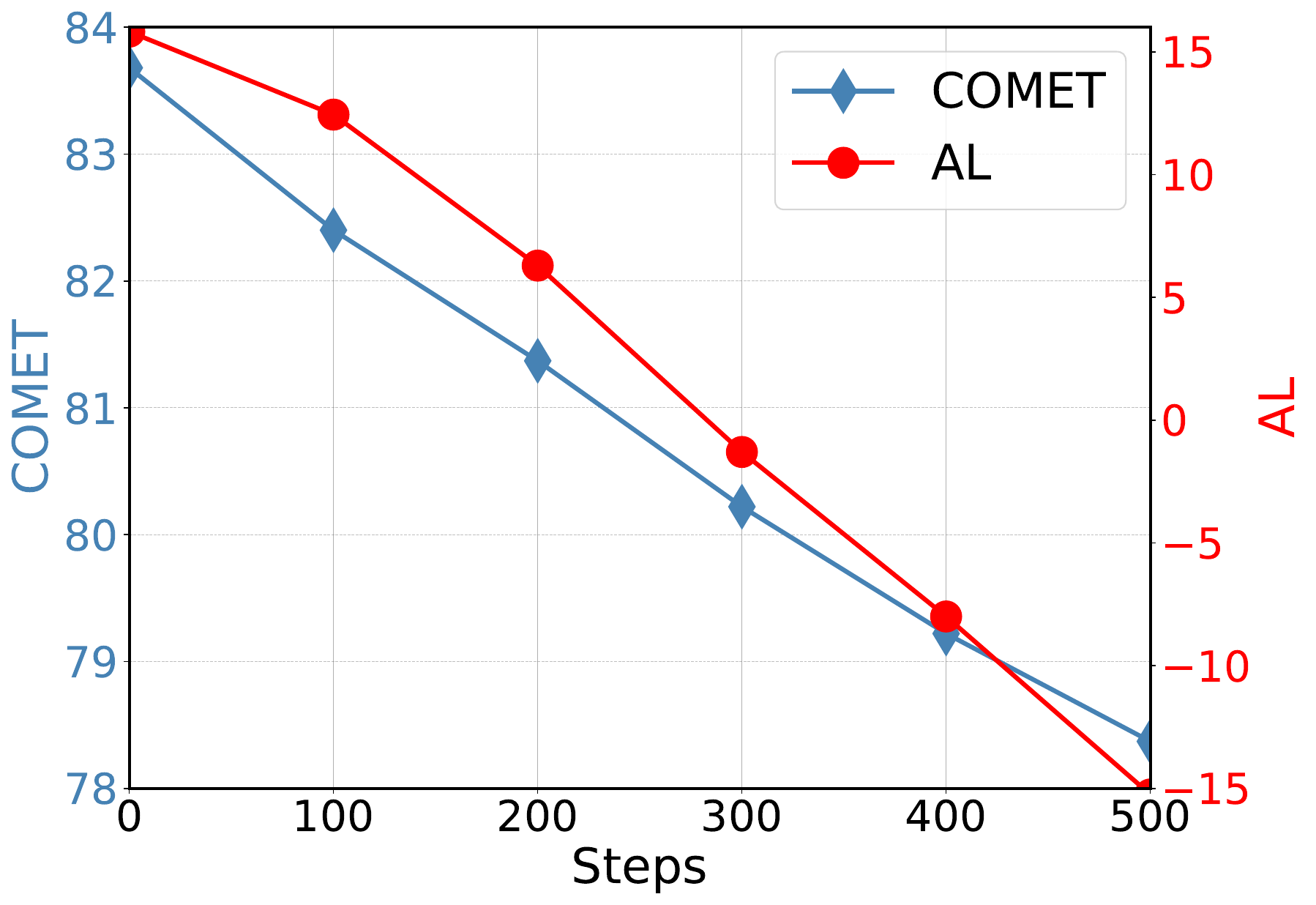}
        \caption{Change of COMET and AL without truncation module during Training.}
        \label{fig:truncation}
    \end{subfigure}
    \caption{Ablation study on the truncation module.}
    \label{fig:ablatation-truncation}
\end{figure}

\paragraph{Simple truncation effectively avoids overfitting to latency.} Our reward design incorporates a carefully designed truncation module. Figure \ref{fig:ablatation-truncation} compares the training dynamics with and without this module. As shown in Figure \ref{fig:truncation}, the absence of the truncation mechanism leads to a sharp decline in AL during model training, even reaching negative values, accompanied by a persistent decrease in the COMET score. Upon inspecting the model's translation output, we observe a tendency to produce numerous meaningless tokens. We hypothesize that this occurs because the model can easily reduce latency by simply outputting such tokens, thereby artificially increasing the overall reward and leading to overfitting to the latency metric. However, as Figure \ref{fig:wotruncation} illustrates, the model with the truncation module effectively avoids AL overfitting while balancing latency and translation quality. These results demonstrate that truncation, despite its inherent simplicity, effectively mitigates overfitting to latency.

\begin{figure}
    \centering
    \includegraphics[width=\linewidth]{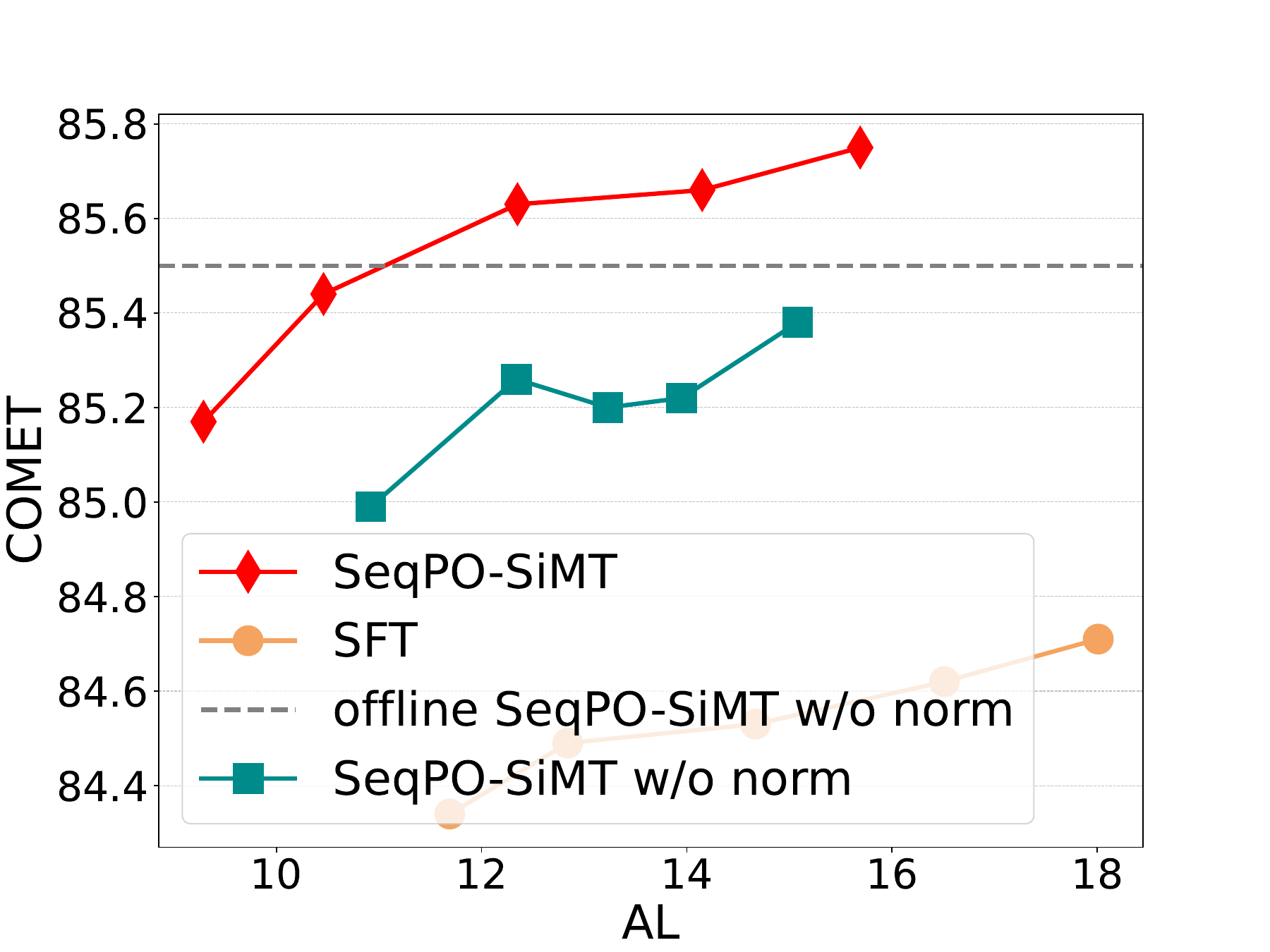}
    \caption{Ablation study on the normalization module.}
    \label{fig:ablation-std}
\end{figure}

\paragraph{Normalization can better balance quality and latency.}
Our method incorporates normalization for both quality and latency scores. Figure \ref{fig:ablation-std} presents the results of the ablation study on the normalization module. As illustrated in the figure, the removal of the normalization module results in a degradation of quality at the same latency levels. These findings demonstrate that normalization, despite its inherent simplicity, effectively balances quality and latency.

\begin{table}[ht]
    \centering
    \small
    \begin{tabular}{lccccc}
      \toprule
      & BLEURT & COMET & GPT-4 \\
      \midrule
      
      & \multicolumn{3}{c}{MUSTC En $\to$ Zh}\\
      offline SFT & \underline{66.28}	& \underline{86.94}&	\underline{92.61}\\
      offline LLaMa3 & 60.43 & 83.78 & 86.98\\
      offline Qwen2.5 & 65.47	&86.49	&91.97\\
      offline \method & \textbf{67.59}	& \textbf{87.74}	& \textbf{93.33}\\
      \midrule
       & \multicolumn{3}{c}{REALSI Zh $\to$ En}\\
      offline SFT  & \underline{65.06}	&\underline{83.92}	&83.72\\
      offline LLaMa3 &  63.27 & 81.19 & 85.66\\
      offline Qwen2.5 & 64.08	 & 82.14	& \underline{85.79}\\
      offline \method & \textbf{66.82}	&\textbf{84.62}	&\textbf{86.79}\\
      \bottomrule
    \end{tabular}
    \caption{Comparison of translation quality between {\method} and other LLMs' offline translation results. The best results and the second-best are highlighted by bold and underscore respectively.}
    \label{tab:appendix-offline}
\end{table}

\paragraph{Additional Results.} Due to space limitations, we have included additional experimental results in the appendix. These results encompass offline translation outcomes, a comparison with traditional encoder-decoder models, an analysis under low-latency conditions, a human evaluation of SiMT results, and a case study.

\section{Related Work}
\label{chap-related}
\subsection{Simultaneous Machine Translation}
Existing SiMT methods can be divided into three categories: rule-based, alignment-based, and reinforcement learning-based. \textbf{Rule-based} methods rely on heuristics. For example, \citet{ma-etal-2019-stacl} proposes a wait-$k$ policy to wait for $k$ tokens before translating. \citet{DBLP:journals/corr/ChoE16} wait until the source input provides more information. \textbf{Alignment-based} methods adapt the full sentence translation model to SiMT by aligning the source sentence and target translation at the word level. For example, \citet{zheng-etal-2019-simpler} convert full sentence translation pairs to partial translation pairs through some heuristic alignment rules, then use the partial data to train the SiMT model. 
\citet{arivazhagan-etal-2019-monotonic}, \citet{communication2023seamlessmultilingualexpressivestreaming}, and \citet{zhang-etal-2024-streamspeech} integrate an alignment module into the model, like monotonic attention and Connectionist Temporal Classification \citep{alex2006ctc} to align the source and target sentence. However, we claim that the alignment process is highly noisy. 
\textbf{Reinforcement learning-based} methods build upon an existing offline translation model by adding a new read/write policy \citep{grissom-ii-etal-2014-dont, satija2016simultaneous, gu-etal-2017-learning, alinejad-etal-2018-prediction, ive-etal-2021-exploiting}. These methods typically treat the translation model as the environment and focus exclusively on optimizing the read/write policy. However, the translation model is only trained on full-sequence translation pairs. They sacrifice a significant amount of translation quality to improve latency, resulting in poor experimental outcomes.

\subsection{Reinforcement Learning from Human Feedback}
RLHF is a technique that aligns LLMs with human preferences. There are many RLHF methods shown to be effective, such as PPO \citep{DBLP:journals/corr/SchulmanWDRK17}), DPO \citep{rafailov2024direct}, and GRPO \citep{shao2024deepseekmath}. RLHF has been successfully applied to various applications, such as ensuring safety \citep{dai2023safe} and mitigating toxicity \citep{korbak2023pretrain}. However, they mainly tackle single-step generation processes while SiMT is a multi-step decision making process.

\section{Conclusion}
In this work, we introduce Sequential Policy Optimization for Simultaneous Machine Translation (SeqPO-SiMT), a novel framework that defines the SiMT task as a sequential decision-making problem. Specifically, we conduct multi-step SiMT data sampling processes and optimize according to quality and latency. This intuitive framework allows the SiMT model to effectively refine the translation quality and reduce latency.  
We conduct extensive experiments on six datasets from the diverse domains for En $\to$ Zh and Zh $\to$ En SiMT tasks. Experimental results demonstrate that {\method} consistently achieves significantly higher translation quality with lower latency. Moreover, {\method} achieves comparable translation quality as high-performing offline translation models, such as Qwen-2.5-7B-Instruct and LLaMA-3-8B-Instruct. 

\section{Limitations}
While this work achieves good performance on 7B LLMs, we cannot ensure that this method can scale to extremely large LLMs, like Deepseek-V3-671B. Future works can scale to larger language models. The current method is still bilingual, and as the number of languages increases, balancing quality and latency across multiple languages presents significant challenges. Future research could expand to multilingual SiMT.

\bibliography{anthology,custom}
\newpage
\clearpage
\appendix
\section{Experimental Settings}
In this section, we offer additional implementation details regarding our experiments, as well as a comprehensive overview of the SFT data construction process involved in this work.
\subsection{Implementation Details}
\label{appendix:exp-setup}
\begin{figure*}
    \centering
    \includegraphics[width=0.7\linewidth]{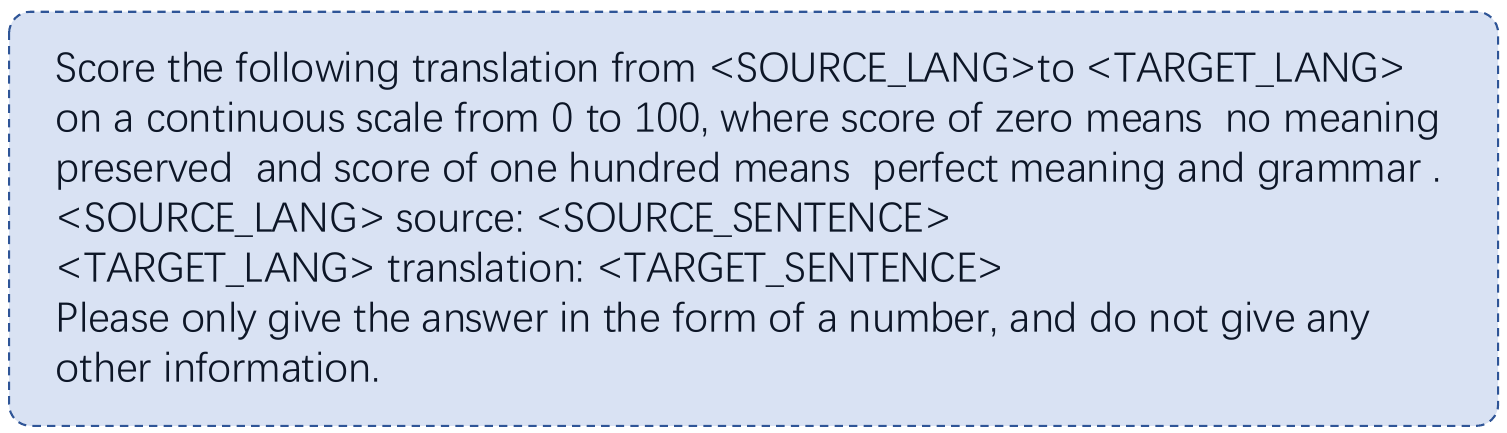}
    \caption{Prompt template while scoring translation results from different models.}
    \label{fig:score-prompt}
\end{figure*}

We use the pretrained model provided by HuggingFace and run all the experiments on NVIDIA A100 GPU with pytorch. And we use the code provided in trl \footnote{https://github.com/huggingface/trl} to train the SFT model. 
The hyper-parameters of the training and generation used in our experiments are shown in Table \ref{tab:hyper-parameter}. During multi-step SiMT sampling, we randomly sample five translations with greedy search and temperature $=1.0$. Figure \ref{fig:score-prompt} illustrates the template used to score translations from different models. Translation results were evaluated using gpt-4o-2024-08-06.
\begin{table}[ht]
    \centering
    \small
    \begin{tabular}{lc}
    \toprule
    Transformer Hyper-parameters \\
    \midrule
        optimizer & AdamW\\
        adam-$\beta$ & (0.9, 0.999)\\
        gradient clipping & 1.0 \\
        gradient accumulation steps & 8\\
        learning rate & $1\times 10^{-6}$\\
        precision & bf16\\
        batch size & 32 \\
    \toprule 
    Generation Hyper-parameters \\
    \midrule
    temperature & 1.0\\
    max new tokens & 60 \\
    top\_k & 0\\
    top\_p & 1.0 \\
    do\_sample & True \\
    \bottomrule
    \end{tabular}
    \caption{Hyper-parameters of the SiMT model.}
    \label{tab:hyper-parameter}
\end{table}
\subsection{SFT Data Construction}
\label{appendix:sft-data}
\begin{figure*}
    \centering
    \includegraphics[width=\linewidth]{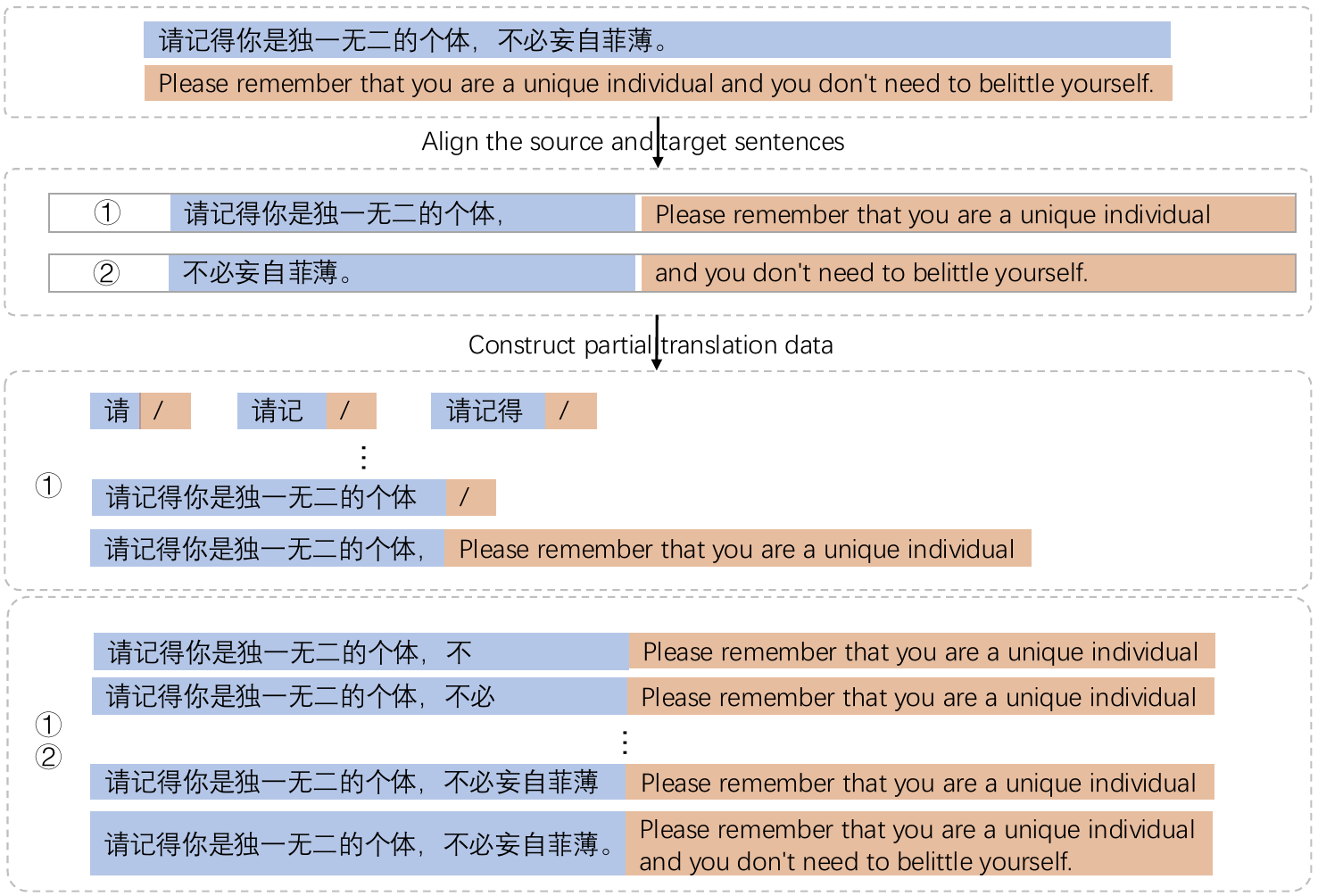}
    \caption{Illustration of SFT data construction process.}
    \label{fig:sft-example}
\end{figure*}

For our SFT data, we randomly sample 40,000 instances from the training datasets. Specifically, we extract 40,000 Zh $\to$ En translation examples from WMT19 and WENET, and another 40,000 En $\to$ Zh translation examples from GigaST. 
We use these full sentence translation pairs to construct partial translation pairs. The process is illustrated in Figure \ref{fig:sft-example}. It begins with aligning the source and target sentences. We prompt LLMs to segment both sentence and target sentences into chunks and pair semantically equivalent text chunks. As shown in Figure \ref{fig:sft-example}, we segment the source and target sentences into four chunks.  Finally, we concatenate the aligned text chunks to generate partial translation data. 

\section{Experimental Results}
In this section, we provide detailed main results, offline \method's results, ablation study on the reward function, and a qualitative case study. 
\subsection{Detailed Main Results}
\label{appendix:main-results}
\begin{figure*}[ht]
    \centering
    \begin{subfigure}[t]{0.32\textwidth}
        \centering
        \includegraphics[width=\linewidth]{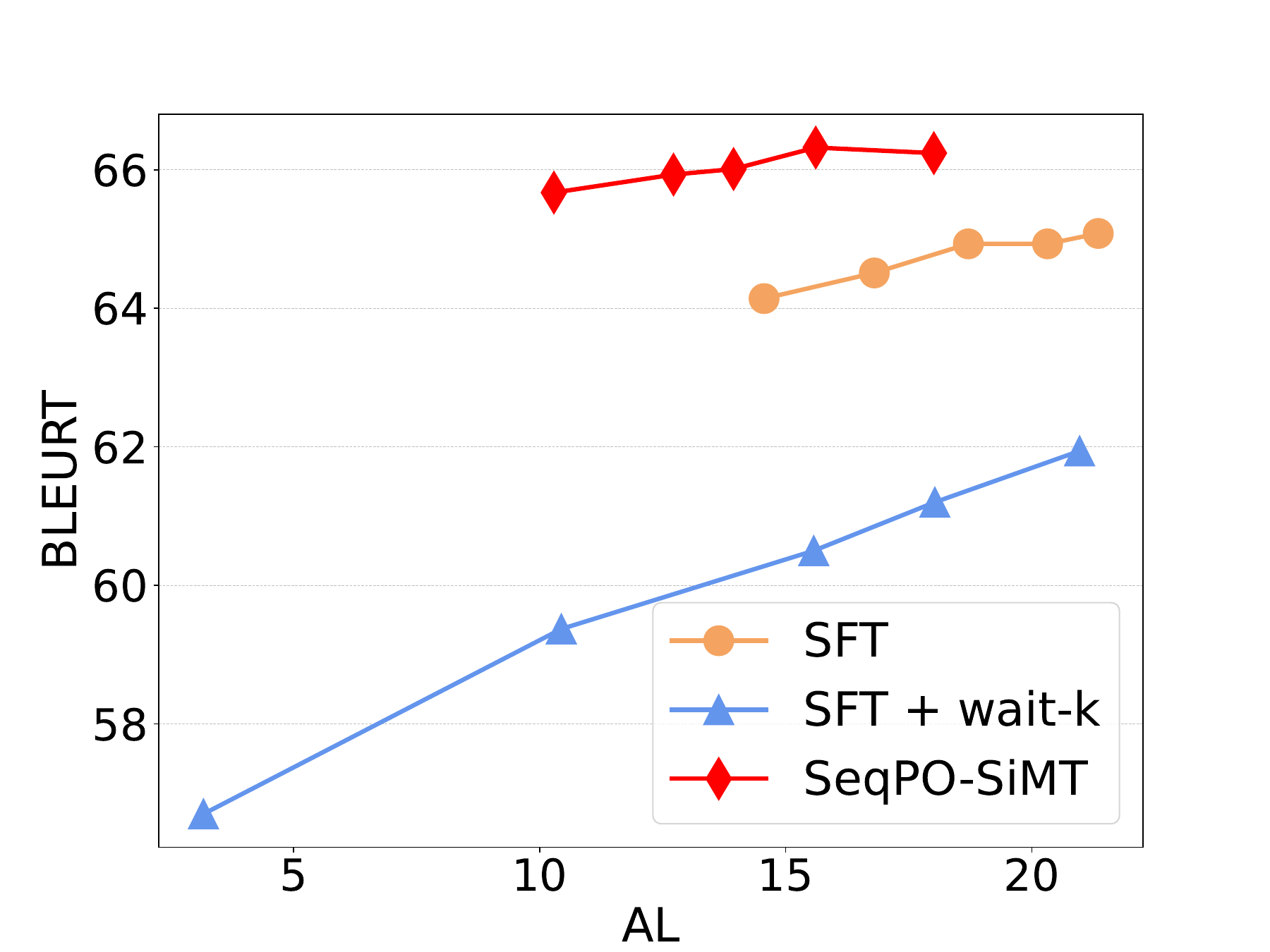}
        \caption{REALSI Zh $\to$ En}
        \label{fig:clasi-zh2en-bleurt-al}
    \end{subfigure}
    \hfill
    \begin{subfigure}[t]{0.32\textwidth}
        \centering
        \includegraphics[width=\linewidth]{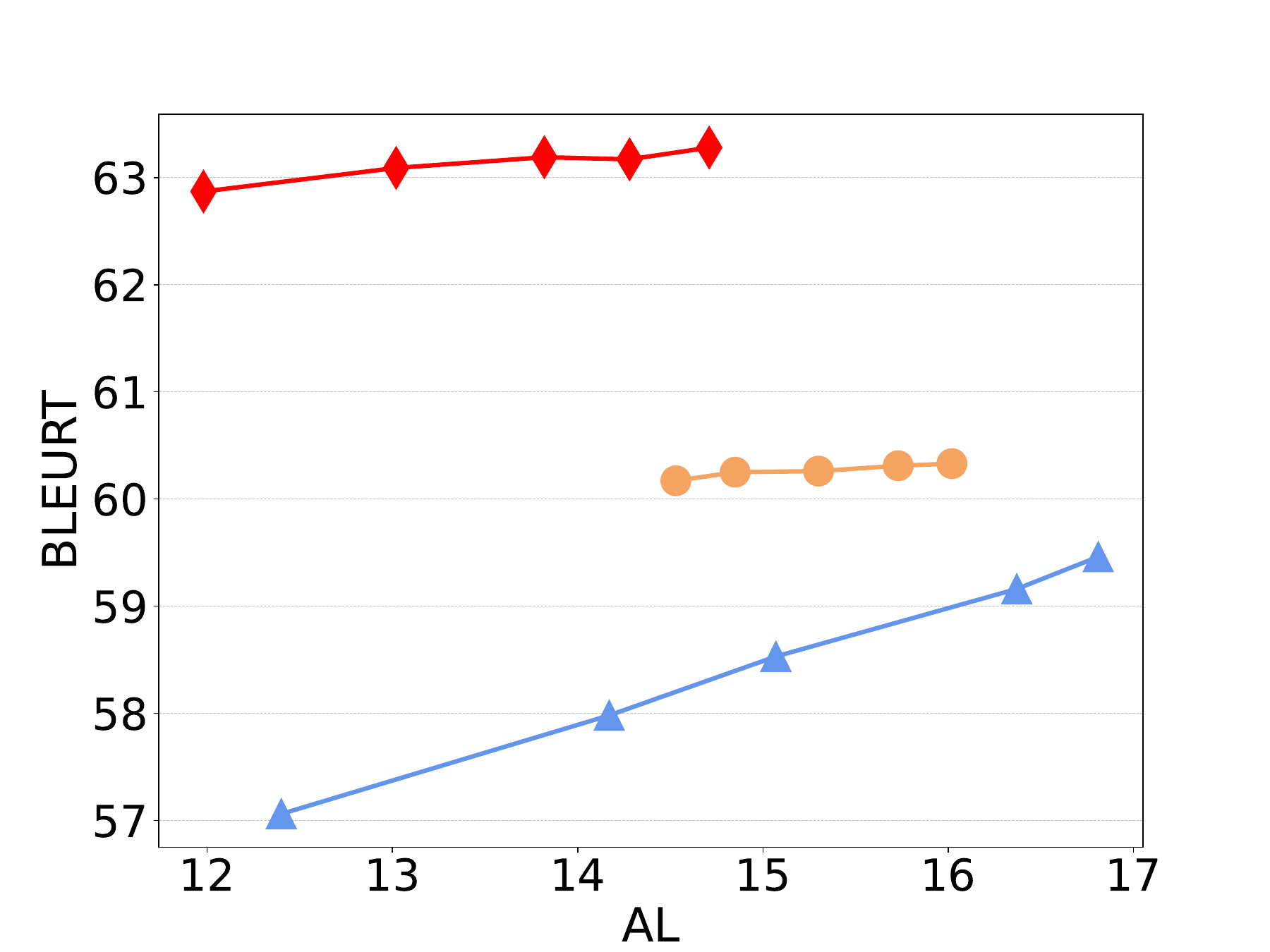}
        \caption{COVOST Zh $\to$ En}
        \label{fig:covost-zh2en-bleurt-al}
    \end{subfigure}
    \hfill 
    \begin{subfigure}[t]{0.32\textwidth}
        \centering
        \includegraphics[width=\linewidth]{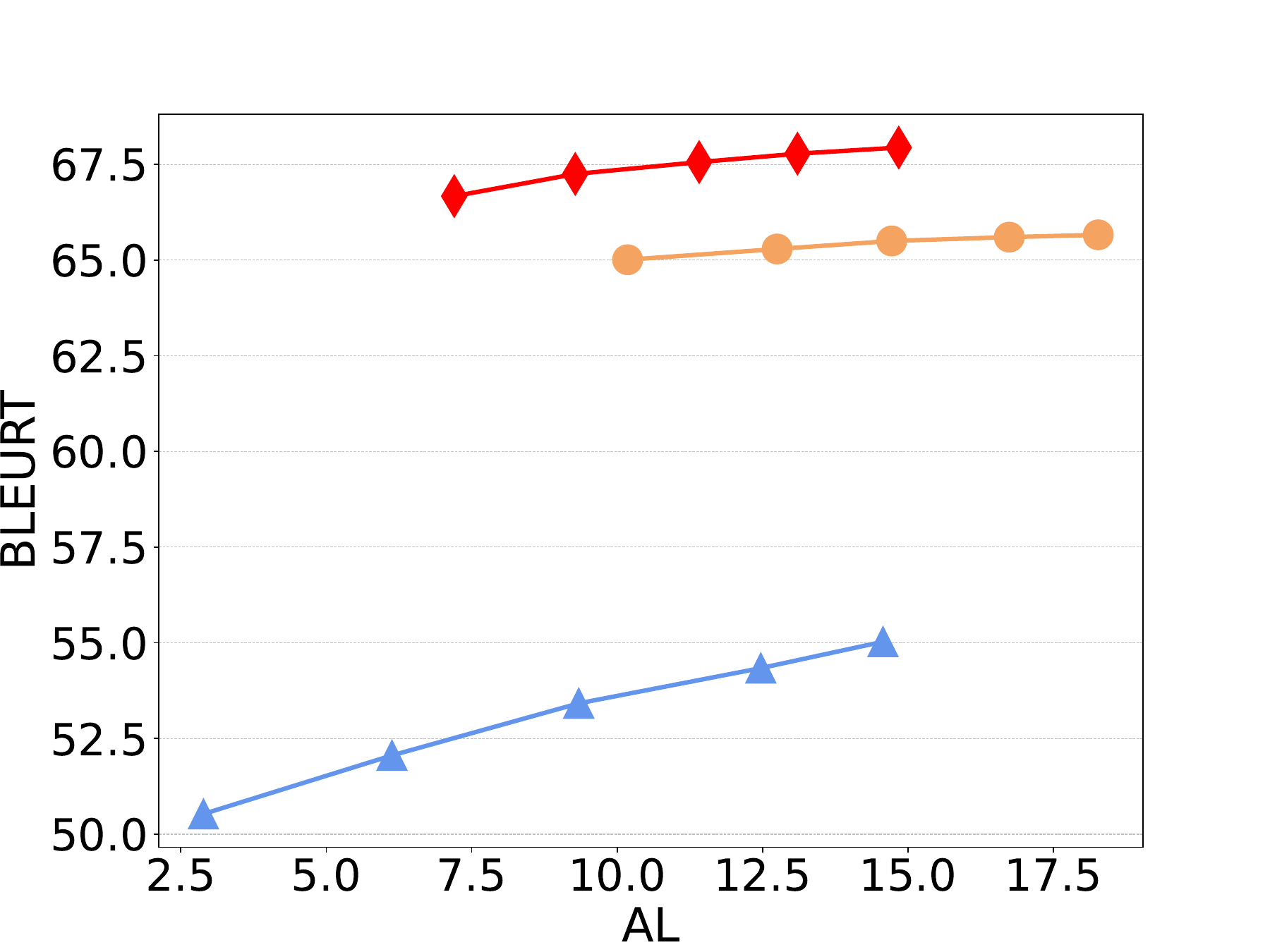}
        \caption{Newstest2021 Zh $\to$ En}
        \label{fig:newstest-zh2en-bleurt-al}
    \end{subfigure}

    \begin{subfigure}[t]{0.32\textwidth}
        \centering
        \includegraphics[width=\linewidth]{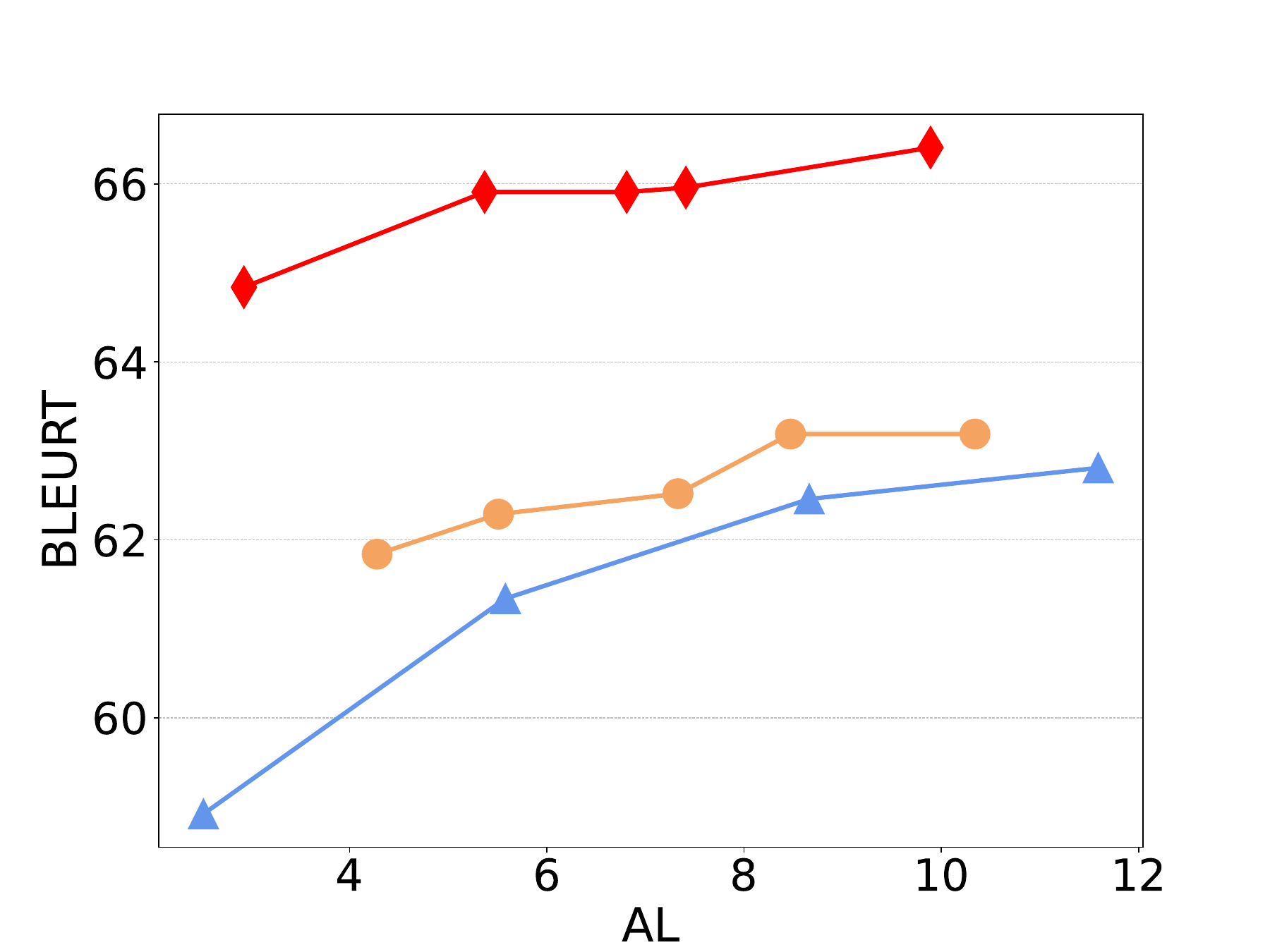}
        \caption{REALSI En $\to$ Zh}
        \label{fig:clasi-en2zh-bleurt-al}
    \end{subfigure}
    \hfill
    \begin{subfigure}[t]{0.32\textwidth}
        \centering
        \includegraphics[width=\linewidth]{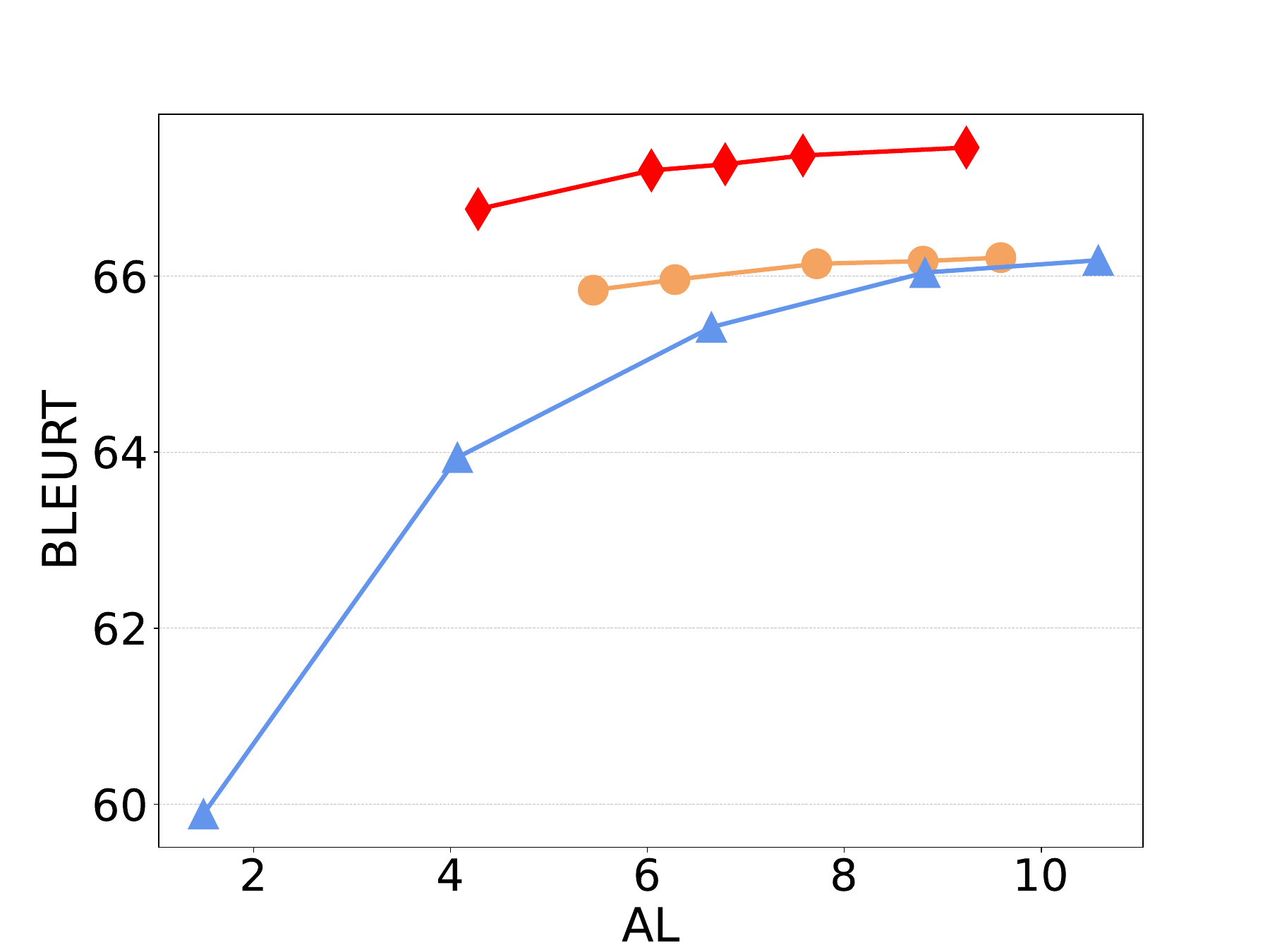}
        \caption{MUSTC En $\to$ Zh}
        \label{fig:mustc-en2zh-bleurt-al}
    \end{subfigure}
    \hfill
    \begin{subfigure}[t]{0.32\textwidth}
        \centering
        \includegraphics[width=\linewidth]{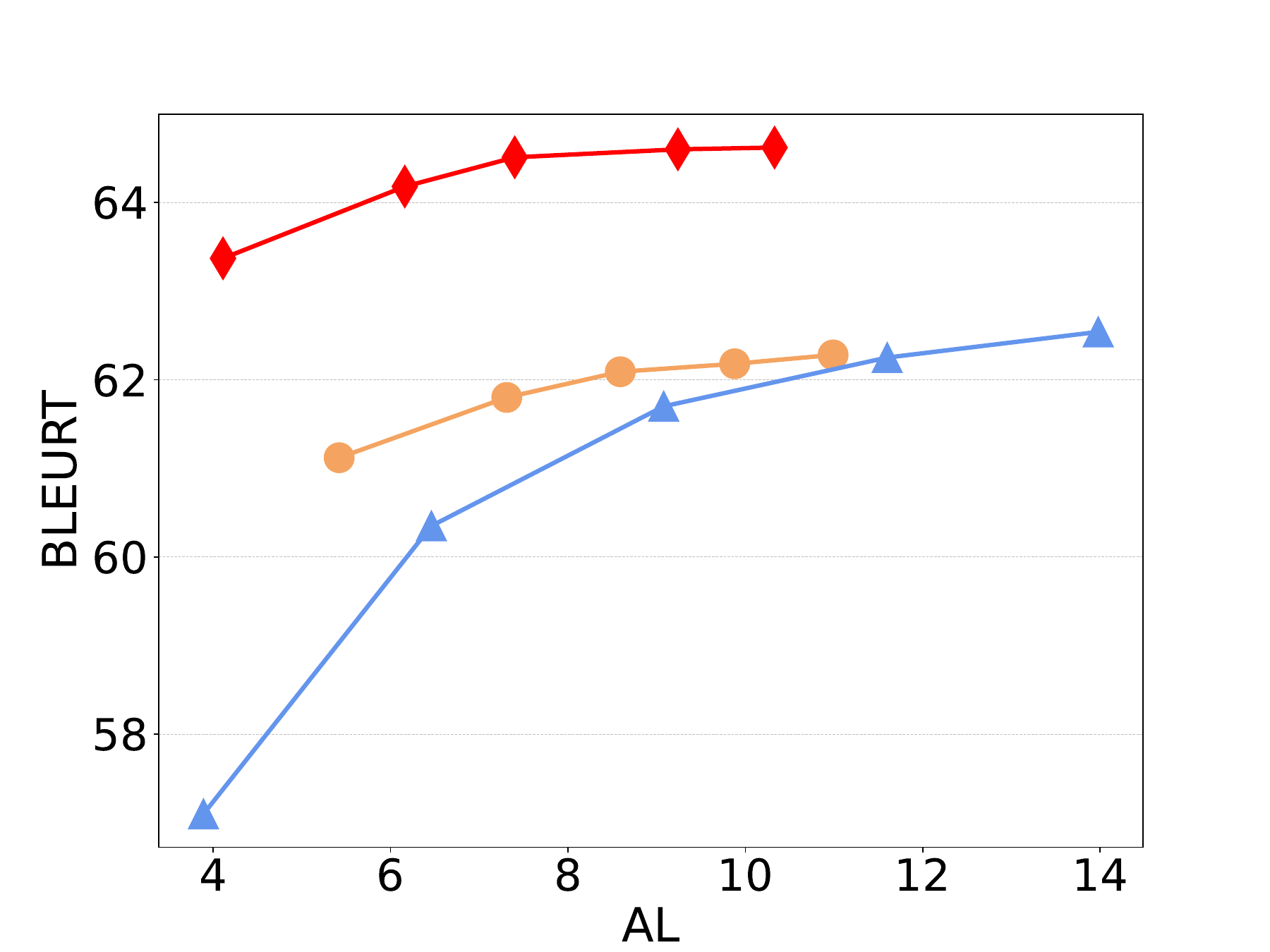}
        \caption{Newstest2021 En $\to$ Zh}
        \label{fig:newstest-en2zh-bleurt-al}
    \end{subfigure}
    \caption{BLEURT against AL on Zh $\to$ En and En $\to$ Zh SiMT tasks. }
    \label{fig:main-bleurt-al}
\end{figure*}

\begin{figure*}[ht]
    \centering
    \begin{subfigure}[t]{0.32\textwidth}
        \centering
        \includegraphics[width=\linewidth]{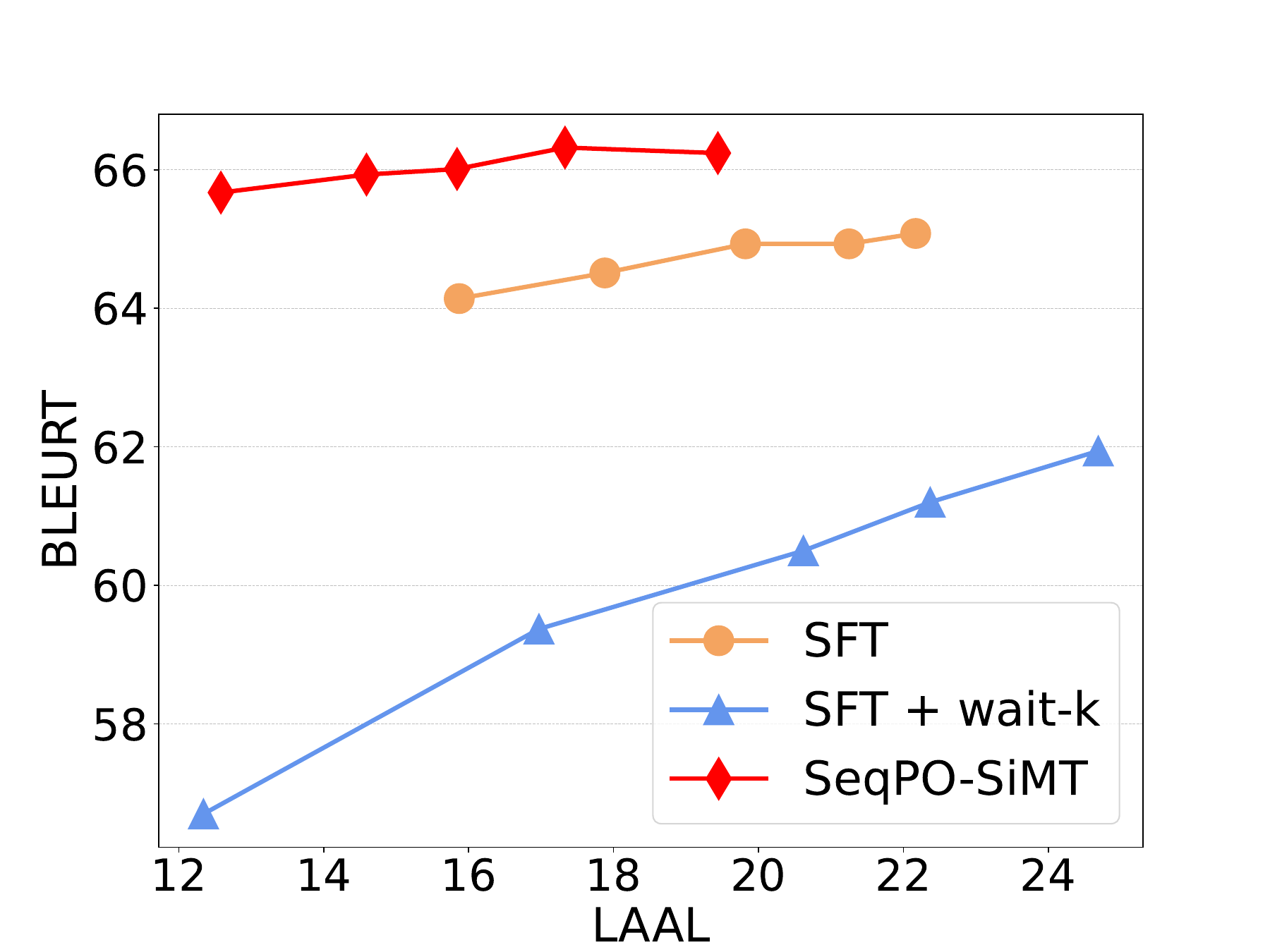}
        \caption{REALSI Zh $\to$ En}
        \label{fig:clasi-zh2en-bleurt-laal}
    \end{subfigure}
    \hfill
    \begin{subfigure}[t]{0.32\textwidth}
        \centering
        \includegraphics[width=\linewidth]{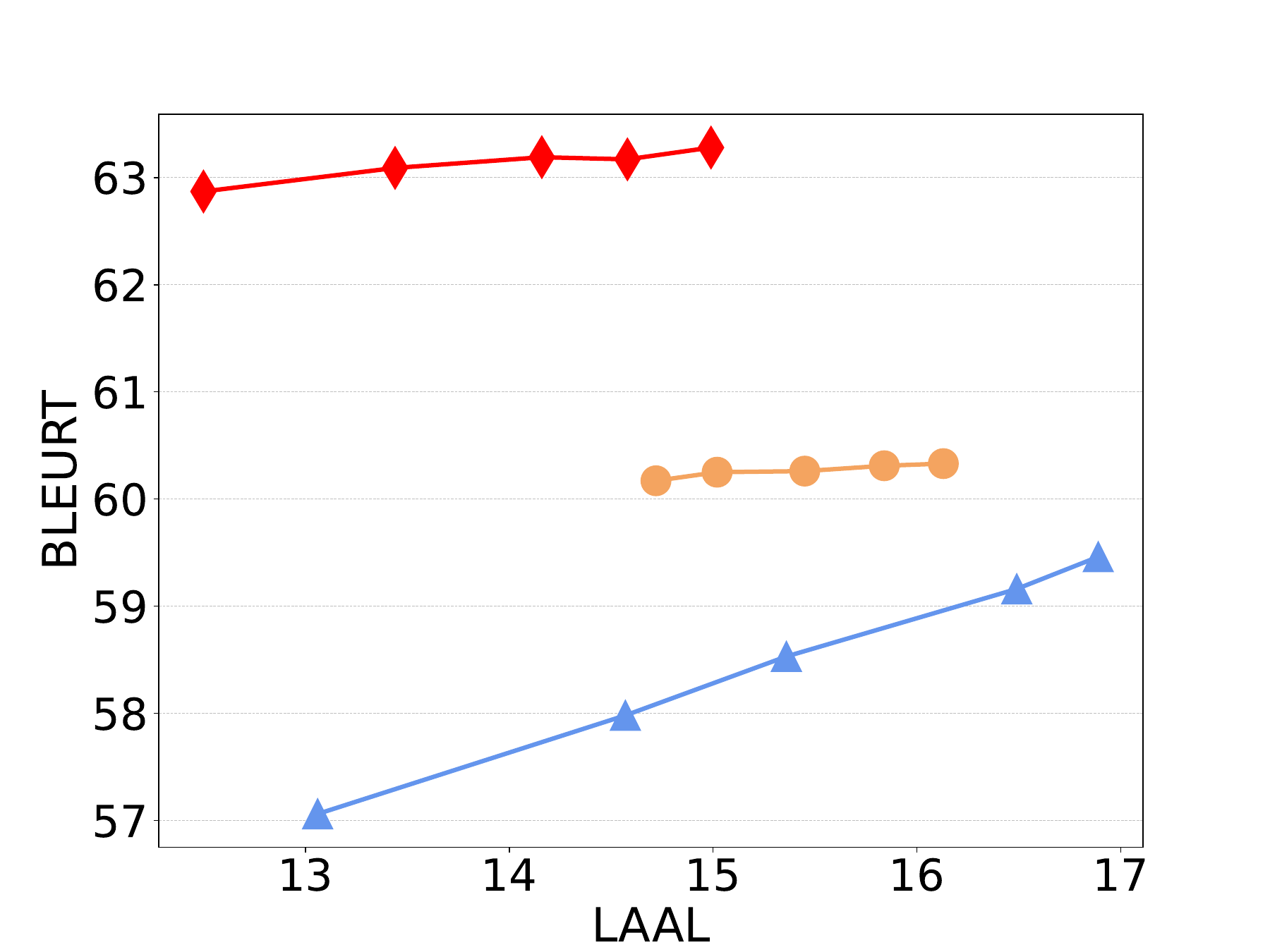}
        \caption{COVOST Zh $\to$ En}
        \label{fig:covost-zh2en-bleurt-laal}
    \end{subfigure}
    \hfill 
    \begin{subfigure}[t]{0.32\textwidth}
        \centering
        \includegraphics[width=\linewidth]{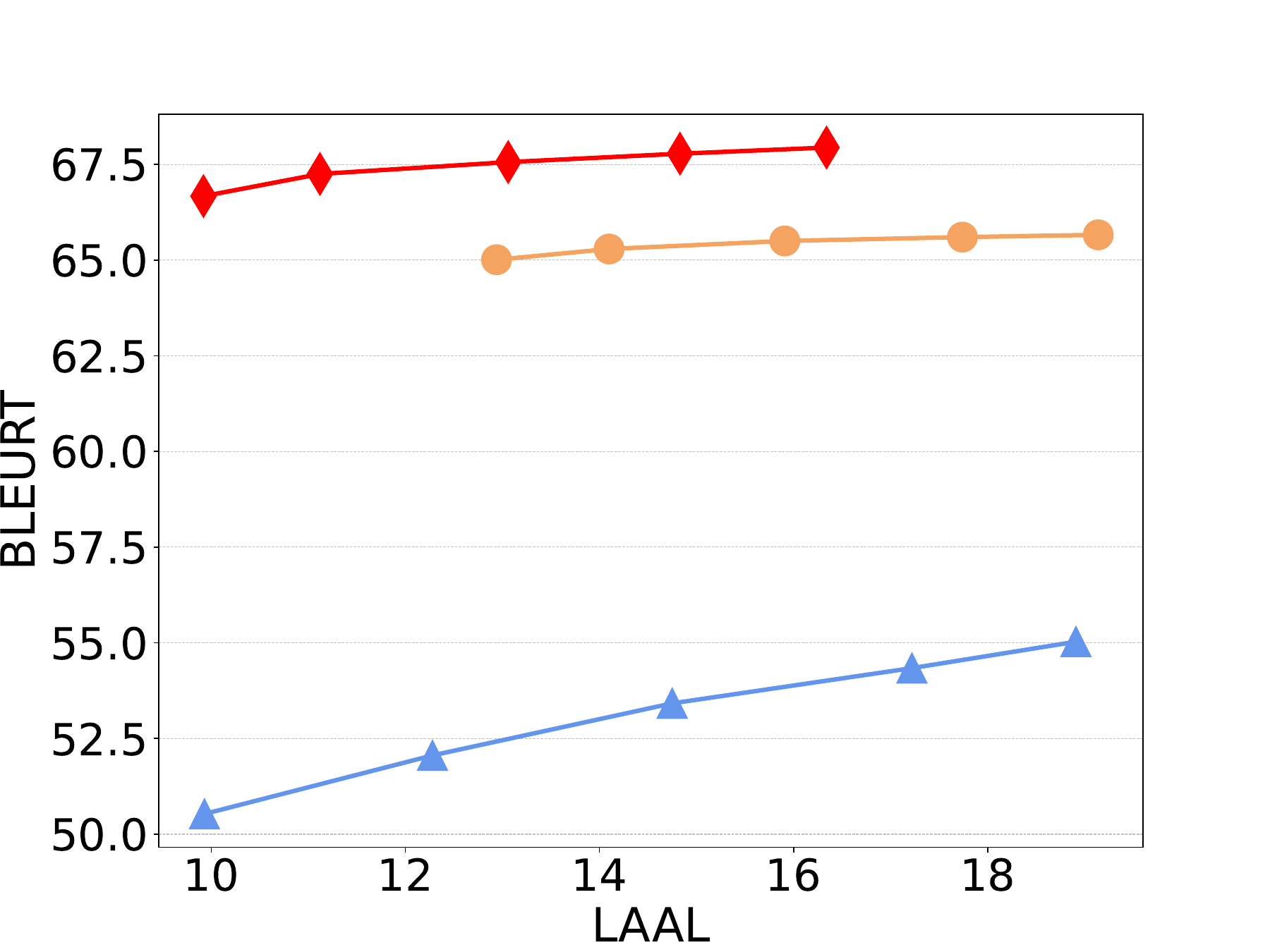}
        \caption{Newstest2021 Zh $\to$ En}
        \label{fig:newstest-zh2en-bleurt-laal}
    \end{subfigure}

    \begin{subfigure}[t]{0.32\textwidth}
        \centering
        \includegraphics[width=\linewidth]{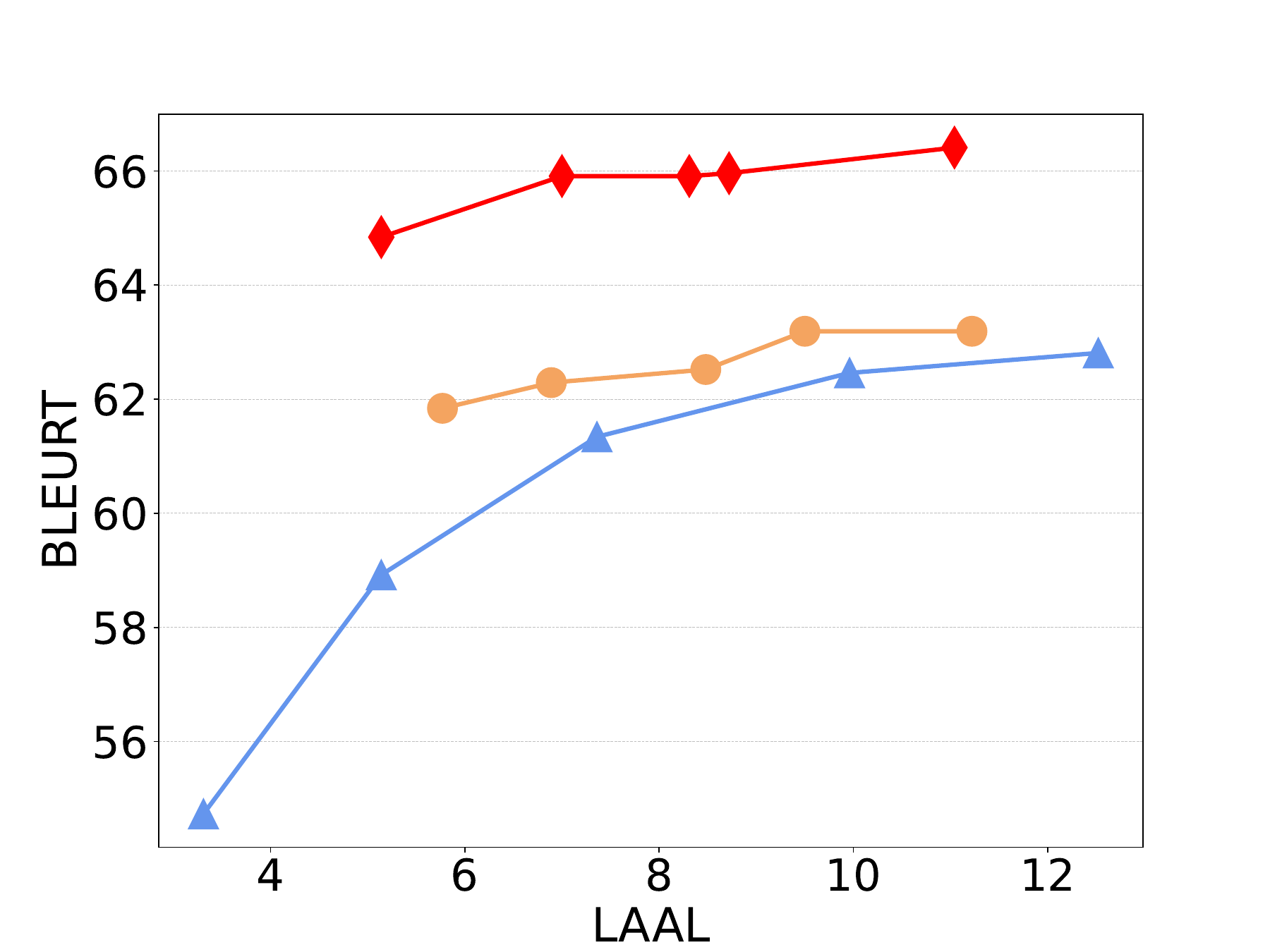}
        \caption{REALSI En $\to$ Zh}
        \label{fig:clasi-en2zh-bleurt-laal}
    \end{subfigure}
    \hfill
    \begin{subfigure}[t]{0.32\textwidth}
        \centering
        \includegraphics[width=\linewidth]{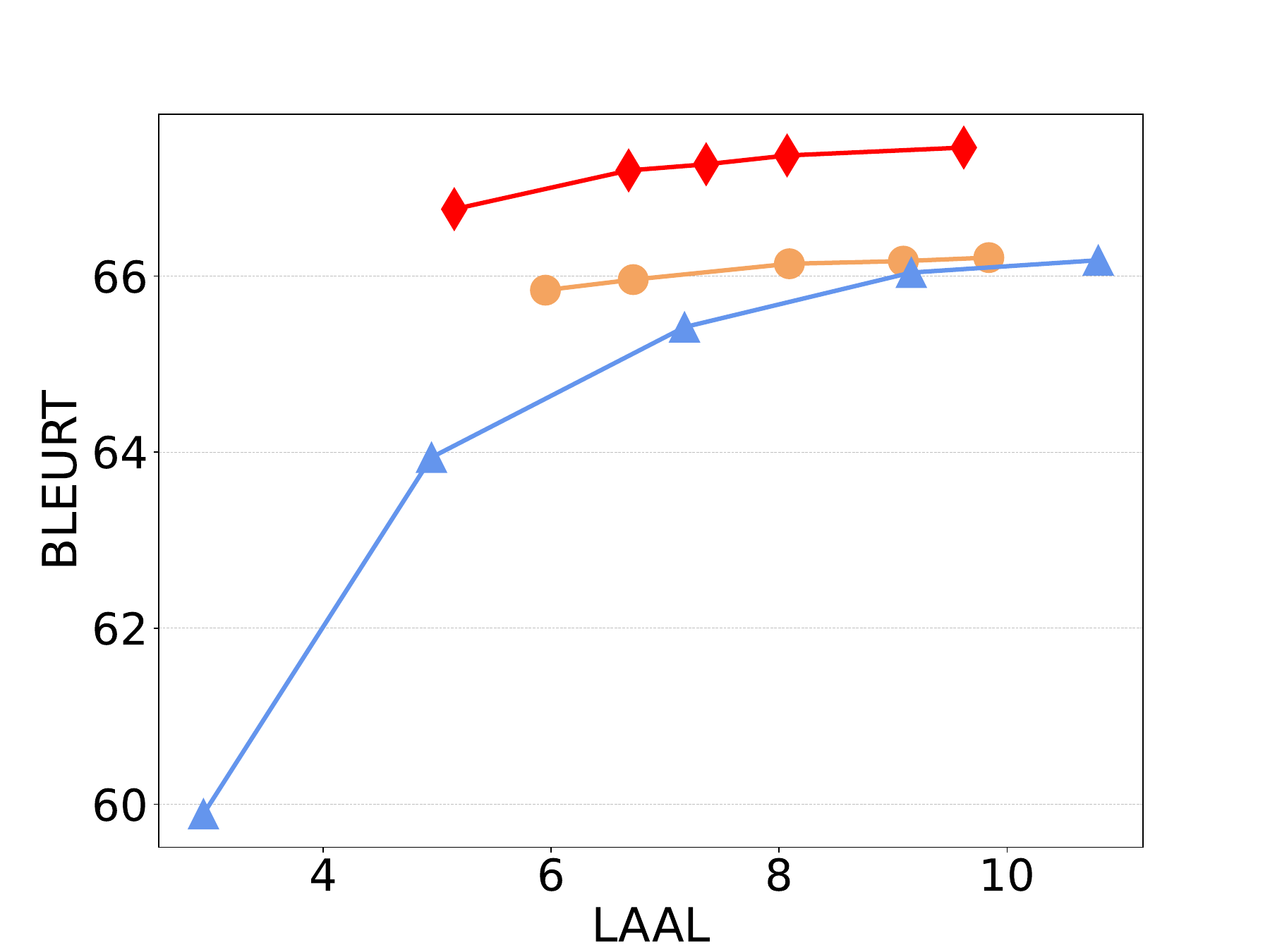}
        \caption{MUSTC En $\to$ Zh}
        \label{fig:mustc-en2zh-bleurt-laal}
    \end{subfigure}
    \hfill
    \begin{subfigure}[t]{0.32\textwidth}
        \centering
        \includegraphics[width=\linewidth]{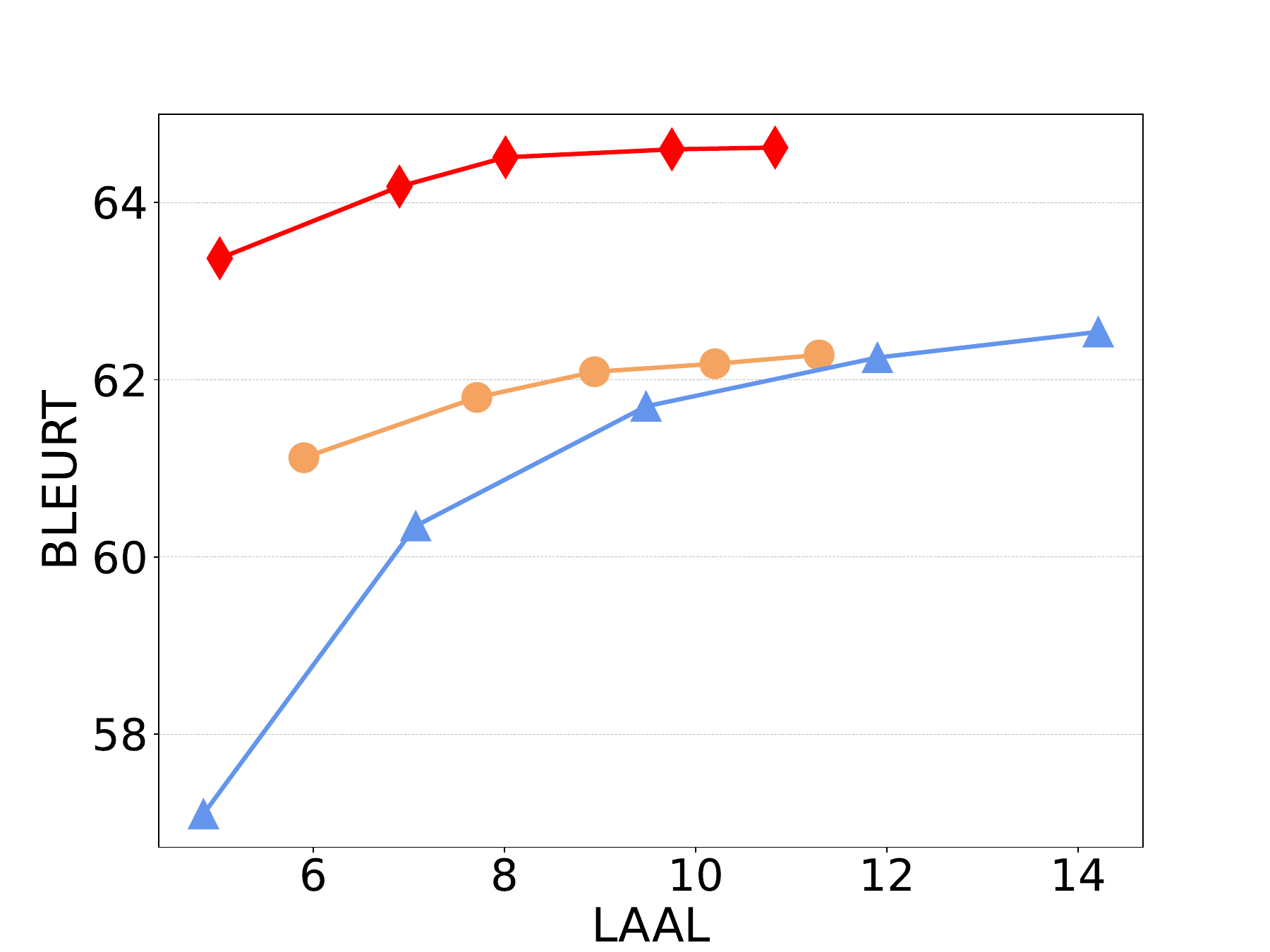}
        \caption{Newstest2021 En $\to$ Zh}
        \label{fig:newstest-en2zh-bleurt-laal}
    \end{subfigure}
    \caption{BLEURT against LAAL on Zh $\to$ En and En $\to$ Zh SiMT tasks. }
    \label{fig:main-bleurt-laal}
\end{figure*}

\begin{figure*}[ht]
    \centering
    \begin{subfigure}[t]{0.32\textwidth}
        \centering
        \includegraphics[width=\linewidth]{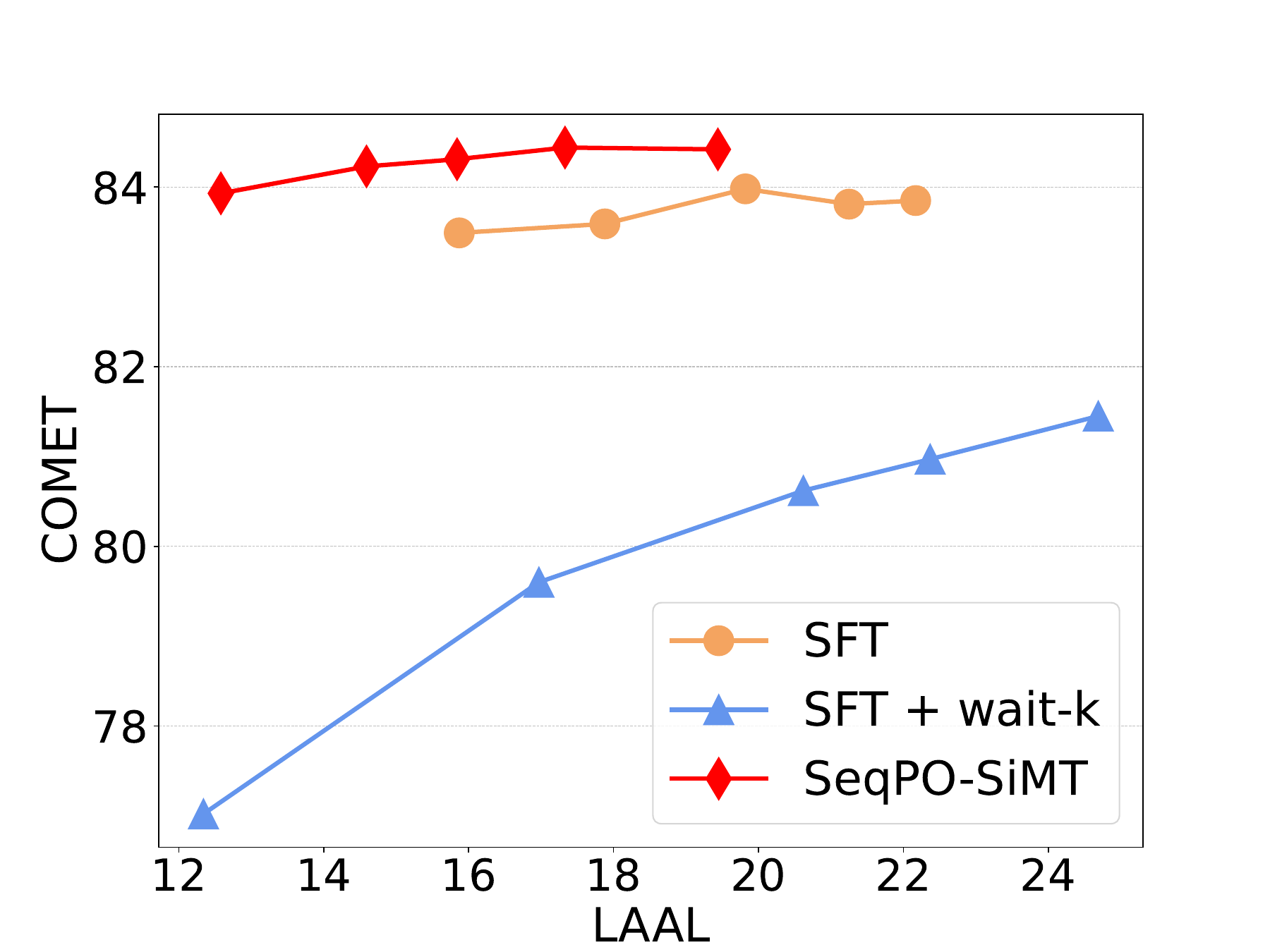}
        \caption{REALSI Zh $\to$ En}
        \label{fig:clasi-zh2en-comet-laal}
    \end{subfigure}
    \hfill
    \begin{subfigure}[t]{0.32\textwidth}
        \centering
        \includegraphics[width=\linewidth]{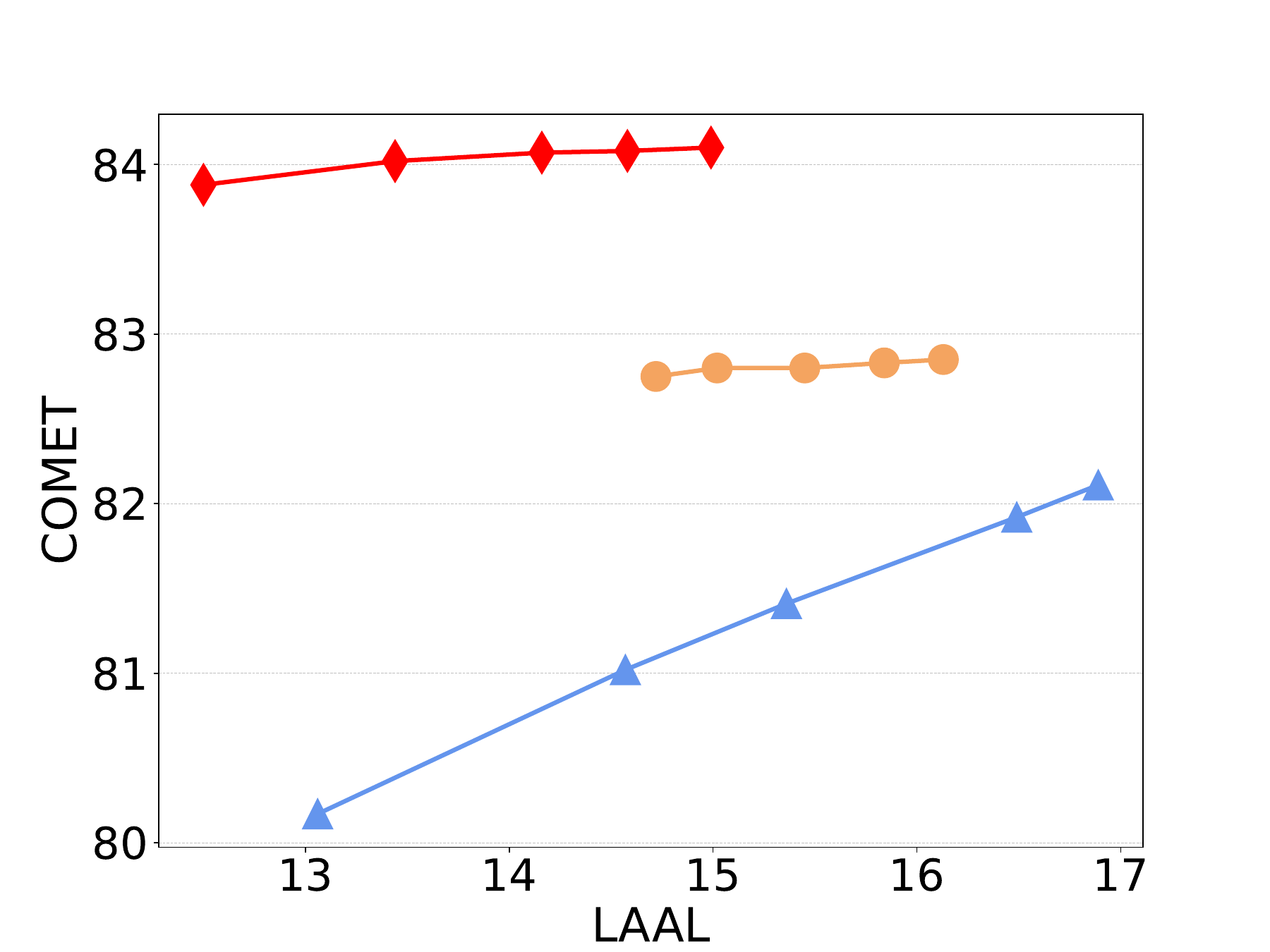}
        \caption{COVOST Zh $\to$ En}
        \label{fig:covost-zh2en-comet-laal}
    \end{subfigure}
    \hfill 
    \begin{subfigure}[t]{0.32\textwidth}
        \centering
        \includegraphics[width=\linewidth]{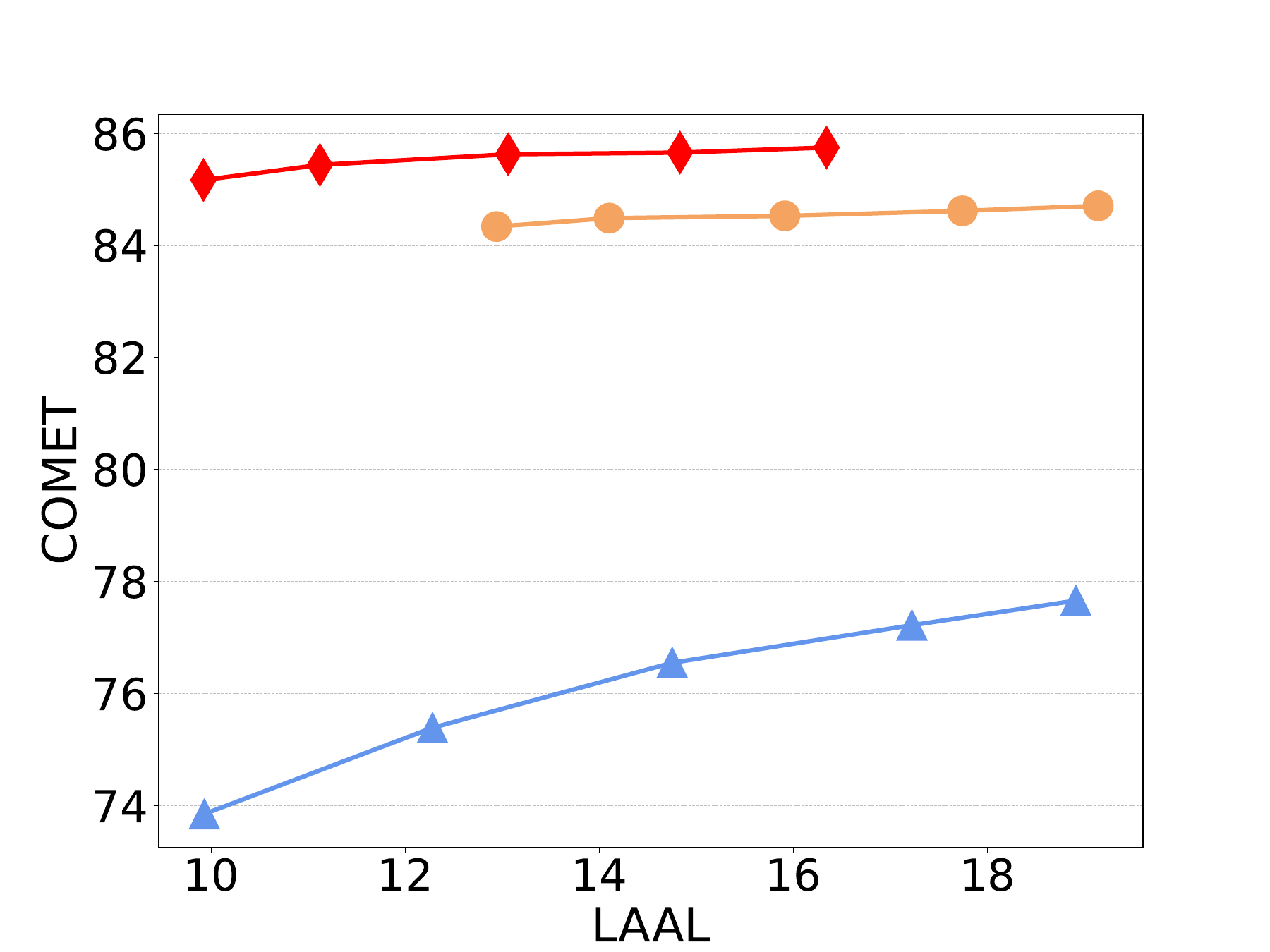}
        \caption{Newstest2021 Zh $\to$ En}
        \label{fig:newstest-zh2en-comet-laal}
    \end{subfigure}

    \begin{subfigure}[t]{0.32\textwidth}
        \centering
        \includegraphics[width=\linewidth]{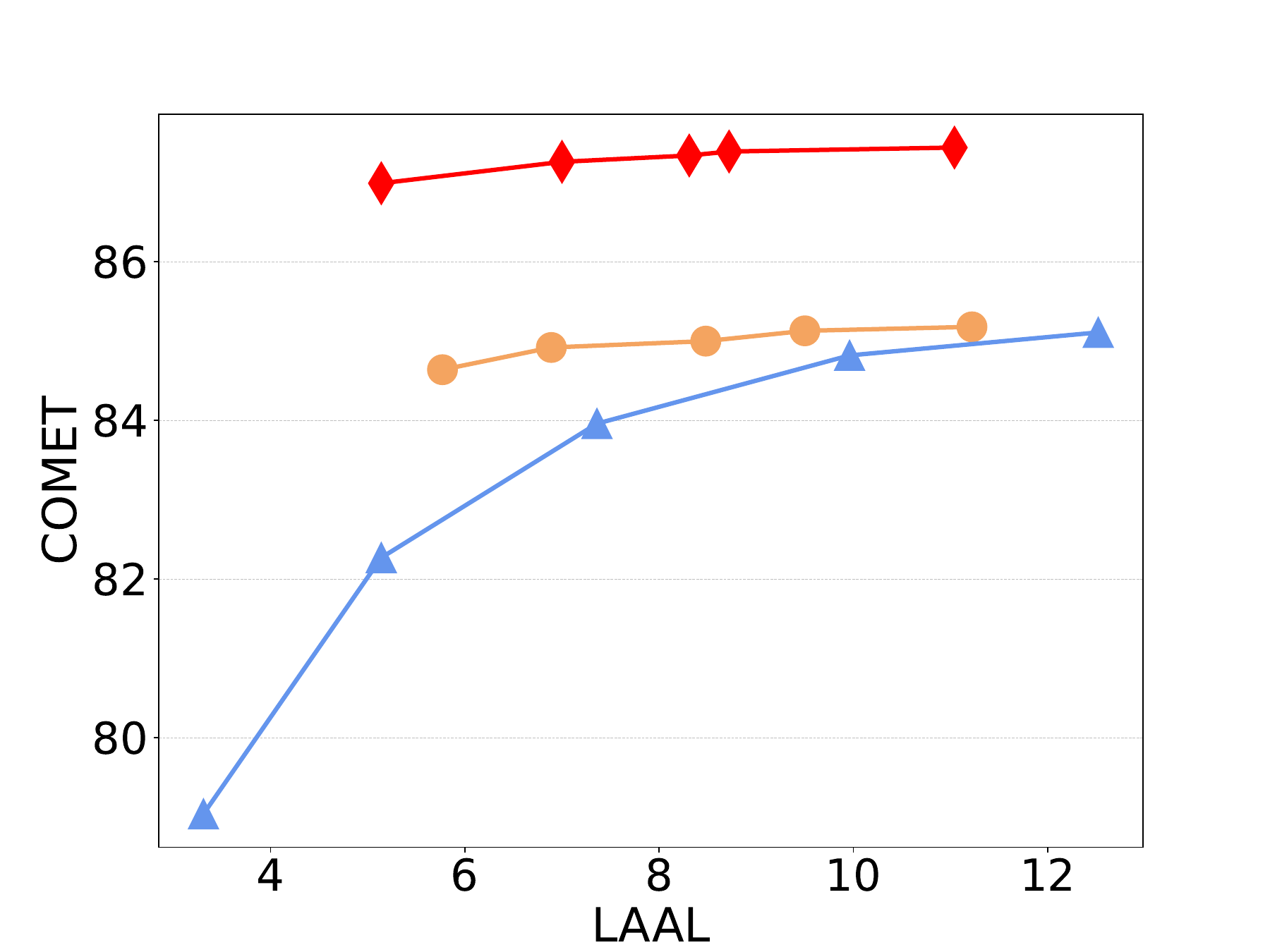}
        \caption{REALSI En $\to$ Zh}
        \label{fig:clasi-en2zh-comet-laal}
    \end{subfigure}
    \hfill
    \begin{subfigure}[t]{0.32\textwidth}
        \centering
        \includegraphics[width=\linewidth]{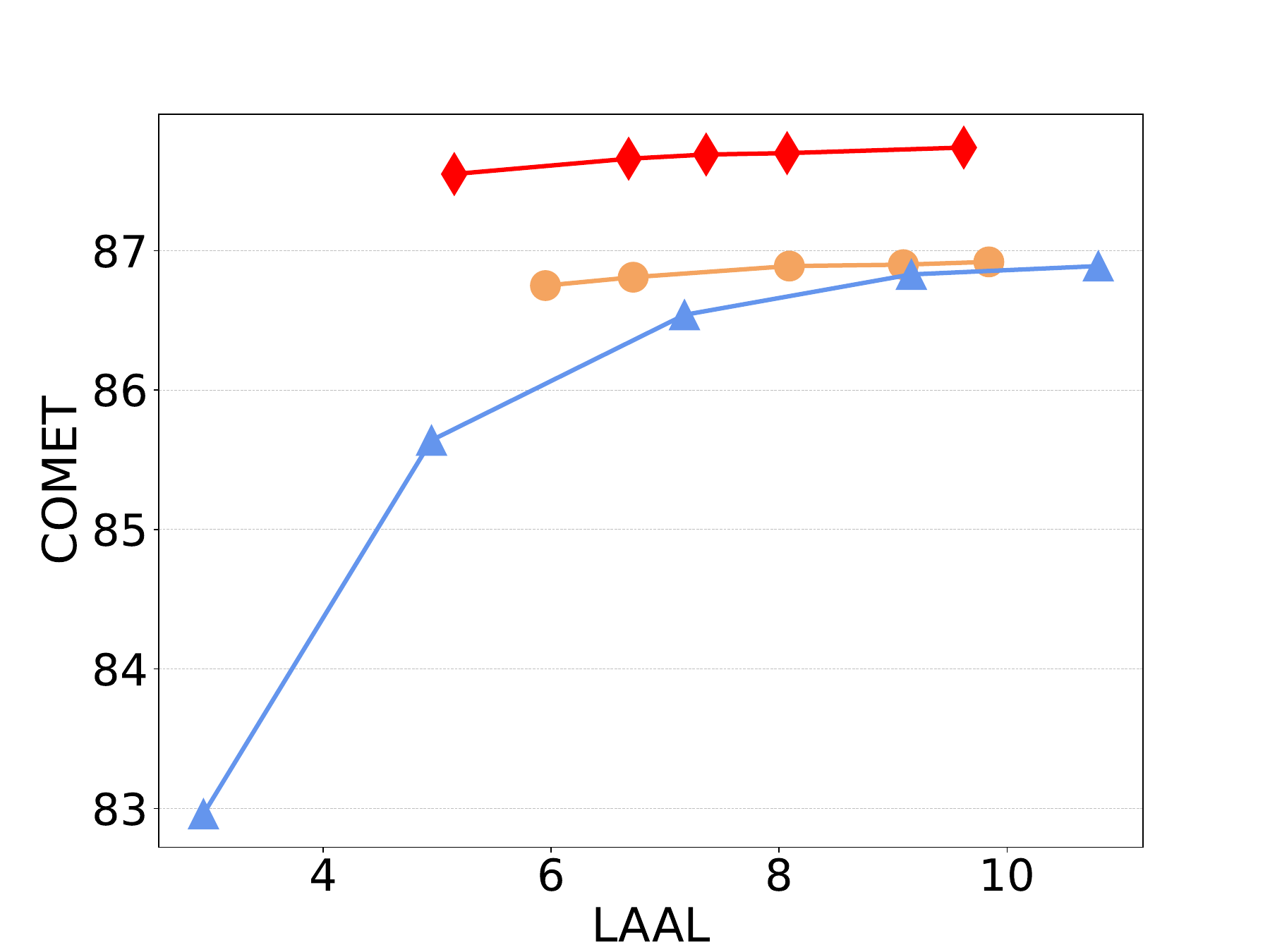}
        \caption{MUSTC En $\to$ Zh}
        \label{fig:mustc-en2zh-comet-laal}
    \end{subfigure}
    \hfill
    \begin{subfigure}[t]{0.32\textwidth}
        \centering
        \includegraphics[width=\linewidth]{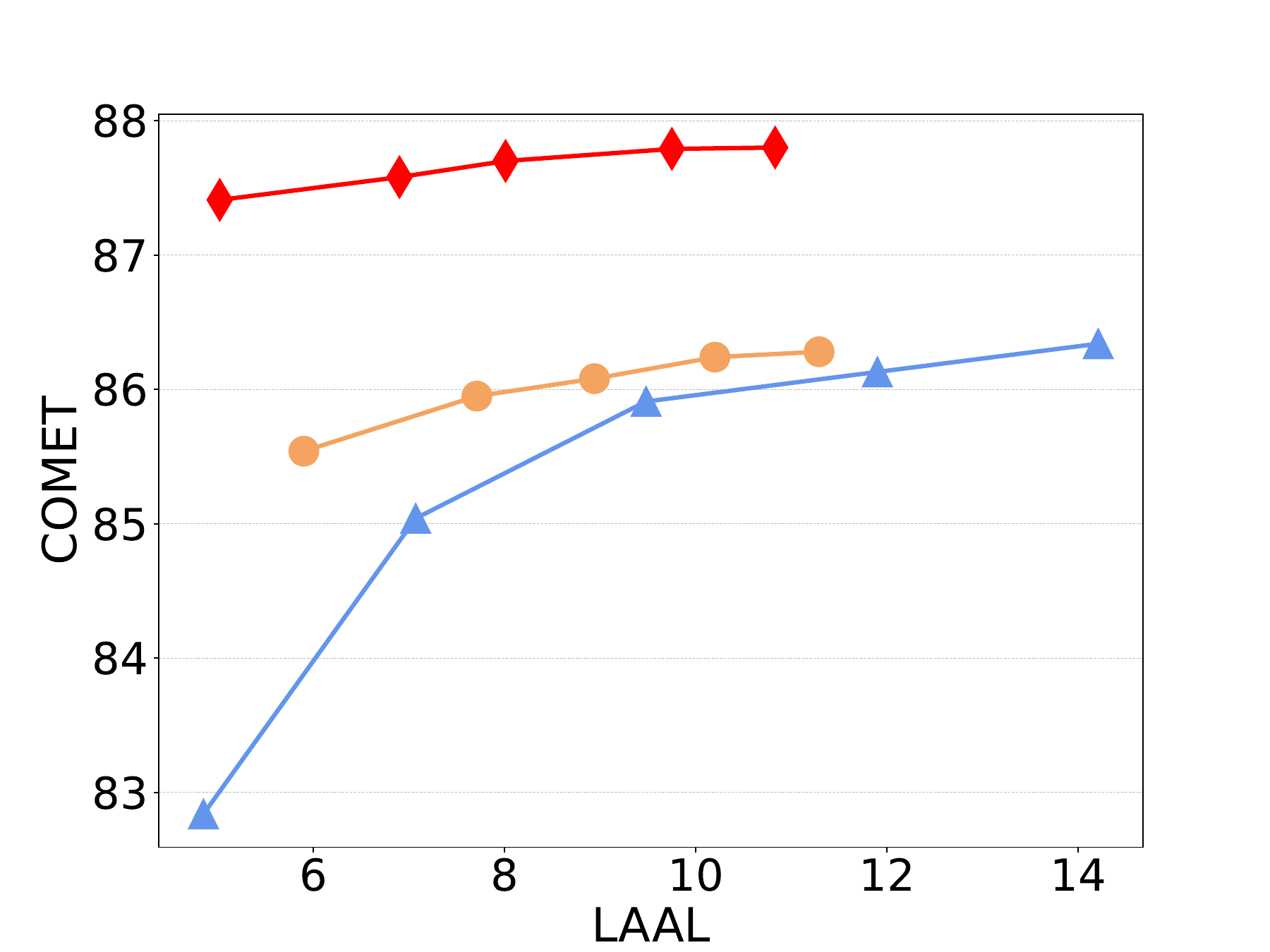}
        \caption{Newstest2021 En $\to$ Zh}
        \label{fig:newstest-en2zh-comet-laal}
    \end{subfigure}
    \caption{COMET against LAAL on Zh $\to$ En and En $\to$ Zh SiMT tasks. }
    \label{fig:main-comet-laal}
\end{figure*}

To comprehensively present the changes in translation quality with respect to translation delay, we have provided the variation graphs of BLEURT v.s. AL, BLEURT v.s. LAAL, and COMET v.s. AL in Figure \ref{fig:main-bleurt-al}, Figure \ref{fig:main-bleurt-laal}, and Figure \ref{fig:main-comet-laal}, respectively. The results show that our proposed {\methodb} achieves a higher translation quality across all latency levels on all datasets, particularly in the low latency level. 

\subsection{Offline \method}
\label{appendix:offline}

We perform offline translation (full sentence translation) using {\method} as base model and compare the results with those of high-performing LLMs in Table \ref{tab:appendix-offline}. The results demonstrate state-of-the-art (SOTA) performance in offline translation.

\subsection{Comparison with Traditional Encoder-Decoder Models}
\label{appendix:cmp-traditional}
We compare the performance of {\method} with traditional encoder-decoder models, including HMT \citep{zhang2023hiddenmarkovtransformersimultaneous}, ITST \citep{zhang-feng-2022-information}, and $\text{SM}^2$ \citep{yu-etal-2024-self}, on NIST 2003-2006 datasets in Table \ref{tab:cmp-traditional}. It is evident that {\method} significantly outperforms other methods across various latency levels. We believe this is due to two main reasons. First, {\method} possesses strong foundational capabilities. Second, {\method} uses both quality and latency as rewards, guiding its reinforcement learning process in a reward-oriented manner, which directly enhances translation quality and reduces translation latency.

\begin{table}[]
    \centering
    \small
    \begin{tabular}{ccccc}
    \toprule
    & \multicolumn{2}{c}{Low Latency} & \multicolumn{2}{c}{High Latency} \\
    \cmidrule(lr){2-3}\cmidrule(lr){4-5}
    & COMET & AL & COMET &AL\\
    \midrule
     HMT    & 78.73 & \textbf{6.11} & 79.85 & \textbf{11.35} \\
    ITST & 79.27 & 6.94 & 79.93 & 11.40 \\
    $\text{SM}^2$ & 79.91 & 6.19 & 80.45 & 11.61\\
    {\method} & \textbf{82.80} & 6.32 & \textbf{83.37} & 11.69\\
    \bottomrule
    \end{tabular}
    \caption{Comparison results with traditional encoder-decoder methods.}
    \label{tab:cmp-traditional}
\end{table}
\subsection{Analysis under very low latency in En $\to$ Zh Setting}
\label{appendix:analysis-low-latency}
Simply using {\method} in the En $\to$ Zh setting does not achieve low latency; however, our method can be combined with traditional read/write policies like wait-k to enable SiMT under low-latency conditions. The results are shown in Table \ref{tab:low-latency}, demonstrating that {\method} significantly outperforms SFT in translation quality at low latency, highlighting the effectiveness of \method.
\begin{table*}[]
    \centering
    \small
    \begin{tabular}{cccccc}
    \toprule
        Dataset & Method & BLEURT  & COMET & AL &LAAL \\
    \midrule
    REALSI     &  SFT + wait-$k$ & 45.73 & 63.35 & 4.08 & 4.11\\
             & {\method} + wait-$k$ & \textbf{55.67} & \textbf{75.98} & \textbf{4.06} & \textbf{4.08}\\
    \midrule
    COVOST  & SFT + wait-$k$ & 45.03 & 65.87 & 3.88 & 3.95 \\
    & {\method} + wait-$k$ & \textbf{55.97} & \textbf{77.69} & \textbf{3.87} & \textbf{3.91} \\
    \midrule
    NEWS & SFT + wait-$k$ & 43.52 & 66.63 & 4.83 & 4.88\\
    &{\method} + wait-$k$ & \textbf{53.53} & \textbf{76.33} & \textbf{4.51} & \textbf{4.59}\\
    \bottomrule
    \end{tabular}
    \caption{Analysis under very low latency in Zh $\to$ En.}
    \label{tab:low-latency}
\end{table*}

\subsection{Human Evaluation}
\label{appendix:human}
\begin{figure}
    \centering
    \includegraphics[width=\linewidth]{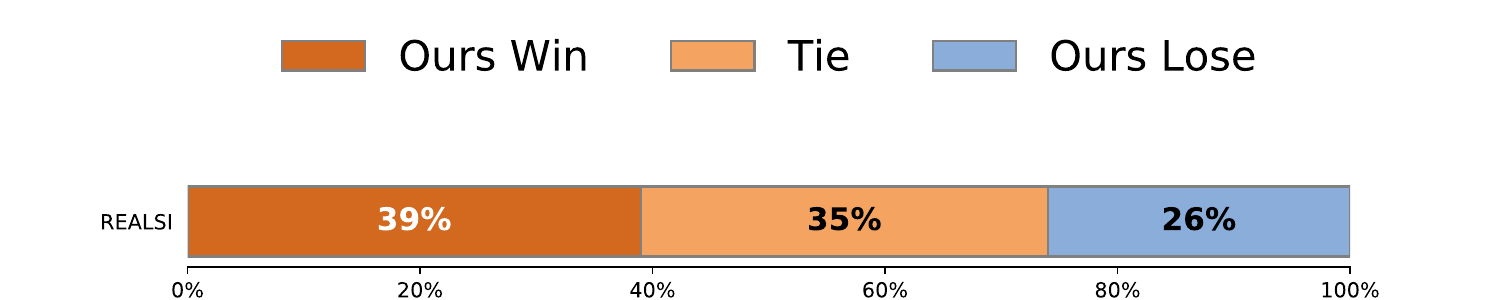}
    \caption{Human Evaluation between {\method} and the SFT model.}
    \label{fig:human}
\end{figure}

To verify that \method aligns with human preference, we randomly sample 100 sentences from the REALSI Zh $\to$ En dataset and conduct a manual evaluation by professional simultaneous interpreters. The results in Figure \ref{fig:human} show that {\method} achieves a
higher win rate than the SFT model, indicating stronger alignment with human preference. 

\subsection{Case Study}
\label{appendix:case}
We analyze the translation results of SFT and {\method} presented in Table \ref{tab:case-study}. Our findings indicate that, given the same source inputs, \methodb exhibits lower latency compared to SFT, allowing it to deliver translations promptly after receiving semantically complete source texts. Furthermore, the SFT model fails to incorporate the information from {\small\begin{CJK}{UTF8}{gbsn}百科辞典\end{CJK}} in the Zh $\to$ En translation and redundantly repeats the phrase \textit{who have been sentenced to do these very harsh sentences} in the En $\to$ Zh translation. In contrast, our method accurately translates the source sentence.
This case study demonstrates that {\method} can achieves higher translation quality alongside lower latency.

\begin{table*}[ht]
    \centering
    \small
    \begin{tabular}{p{1.5cm}p{3cm}p{5cm}p{5cm}}
    \toprule
    Language pair & Source texts & SFT output & \methodb output\\
    \midrule
     \multirow{4}{*}{Zh $\to$ En}&{\small\begin{CJK}{UTF8}{gbsn}该百科辞典\end{CJK}}    &  /&/   \\
     \cmidrule{2-4}
     & {\small\begin{CJK}{UTF8}{gbsn}有电子版和\end{CJK}}    & / & / \\
     \cmidrule{2-4}
     & {\small\begin{CJK}{UTF8}{gbsn}免费的网络\end{CJK}}   & / & This encyclopedia has both a digital version\\
      \cmidrule{2-4}
     &{\small\begin{CJK}{UTF8}{gbsn}版。\end{CJK}} & there is also a digital version and a free online version. & and a free online version.\\
     \toprule
     \multirow{6}{*}{En $\to$ Zh}&And I've been & / & /\\
     \cmidrule{2-4}
     & representing these kids & / & /\\
     \cmidrule{2-4}
     & who have been & {\small\begin{CJK}{UTF8}{gbsn}我一直在代表那些\end{CJK}} & {\small\begin{CJK}{UTF8}{gbsn}我一直在代表这些孩子，\end{CJK}} \\
     \cmidrule{2-4}
     & sentenced to do & {\small\begin{CJK}{UTF8}{gbsn}被判处\end{CJK}} & /\\
     \cmidrule{2-4}
     & these very harsh & / & /\\
     \cmidrule{2-4}
     & sentences. &{\small\begin{CJK}{UTF8}{gbsn}非常严厉刑罚的孩子们。他们被判处了这些非常严厉的刑罚。\end{CJK}}&{\small\begin{CJK}{UTF8}{gbsn}他们被判处了非常严厉的刑罚。\end{CJK}}\\
     \bottomrule
    \end{tabular}
    \caption{Case Study of SFT and \method. A forward slash (/) indicates an empty output. }
    \label{tab:case-study}
\end{table*}

\end{document}